\documentclass[runningheads]{llncs}
\usepackage{graphicx}
\usepackage{amsmath,amssymb} 
\usepackage{color}

\usepackage{algorithm} 

\usepackage[width=122mm,left=12mm,paperwidth=146mm,height=193mm,top=12mm,paperheight=217mm]{geometry}
\begin{document}


\pagestyle{headings}
\mainmatter
\def\ECCV16SubNumber{1729}  

\title{Modeling Visual Compatibility through Hierarchical Mid-level Elements} 

\titlerunning{Modeling Visual Compatibility via Mid-level Elements}
\authorrunning{Oramas \& Tuytelaars}


\author{Jose Oramas M., Tinne Tuytelaars}
\institute{KU Leuven, ESAT-PSI, iMinds}

\maketitle

\begin{abstract}
In this paper we present a hierarchical method to discover 
mid-level elements with the objective of modeling visual 
compatibility between objects. At the base-level, our method 
identifies patterns of CNN activations with the aim of modeling 
different variations/styles in which objects of the classes of 
interest may occur. At the top-level, the proposed method 
discovers patterns of co-occurring activations of base-level 
elements that define visual compatibility between pairs of 
object classes. Experiments on the massive Amazon 
dataset show the strength of our method at describing object 
classes and the characteristics that drive the compatibility 
between them.

\end{abstract}

\section{Introduction}
\label{sec:intro}

With the popularity and high proliferation of camera-capable 
mobile computing devices, the amount of images being collected 
and uploaded to the Internet has increased exponentially in 
recent years. These images cover different types of content, 
e.g. events, locations, consumer products, that are of interest 
to the user. 
Given an image with a content of interest, a common problem is 
to find or retrieve images with related content. The solution 
to this problem has several applications. For tourism, given 
a set of images taken by the user on a previous holiday, 
visual similarity can be used to find and suggest alternative 
destinations where the user may find similar settings than the 
ones of his interest. For shopping, pictures of a specific 
product, e.g. a car, can be used to find accessories that are 
not only adequate for that car model but also that match its 
style. Likewise, given an image of a specific piece of clothes, 
e.g. a shirt,  pants and shoes matching the style of the 
presented shirt can be suggested. 

Over the years, this retrieval/recommendation problem 
has been approached mostly from a text-oriented perspective: 
compatibility between the query image and the 
set of images is determined by analyzing text captions or other
types of metadata associated to the images. This procedure has the weakness 
of not being applicable to images where such text captions or
metadata are not available. This weakness is usually referred to 
in the literature as the ``cold start'' problem 
\cite{Schein2002SIGIR,Zhou2011SIGIR}. Different from 
images published by online shops on the Internet where 
detailed text descriptions are usually available, images taken 
or uploaded by users are not necessarily accompanied 
by such text captions. This makes the 'cold start' problem 
relevant for the proposed applications. To address this 
problem, we propose to do the analysis at the visual level, 
that is, by comparing visual features between the content 
of the images. 

In this work we focus on measuring the visual compatibility 
of objects depicted in different images. Inspired by the 
success of representations based of mid-level elements \cite{doersch2013mid,doersch2014context,doersch2015unsupervised,doersch2015makes,jainCVPR13vidrep,LiLSH15,RematasCVPR15,Singh2012DiscPat}, 
we propose a hierarchical method to learn visual representations 
in a data-driven fashion. At the base-level we discover a set 
of informative class-specific elements by mining activations 
of Convolutional Neural Networks (CNN). These base-level elements 
are effective at describing images of the classes of interest. 
At the top-level we exploit co-occurrences of base-level elements 
between images of compatible objects. This produces a set of 
``rules'' which ``explain'' the compatibility of such objects.
We conduct experiments on the clothes-related classes from 
the massive Amazon-based dataset collected in \cite{McAuleyKDD15}.
We focus on the compatibility estimation problem assuming the 
classes of the objects depicted in the images are known. 
This is a reasonable assumption given the recent impressive 
advances that have been achieved in the image classification 
task \cite{Krizhevsky12}. Moreover, there are significant efforts 
focusing on the task of clothes segmentation \cite{KalantidisClothesParsing2013,SimoSerraACCV2014,SimoCVPR15,YamaguchiKOB15,Yang_2014_CVPR} which can benefit the proposed method.

The main contributions of this work are: i) a novel hierarchical 
method that not only estimates compatibility between objects 
depicted in images but also provides an insight on the features 
that drive the compatibility, and ii) a modular method that is highly 
parallelizable, therefore, it can be trained in a relatively short 
period of time.

This paper is organized as follows: 
in Section~\ref{sec:relatedWork} we position our method
wrt. existing work. In Section~\ref{sec:background} we give a brief 
introduction to Association-Rule Mining for completeness.
Section~\ref{sec:proposedMethod} goes 
into the details of the proposed method to model object 
appearance and visual compatibility while Section~\ref{sec:implementationDetails} 
provides implementation details. In Section~\ref{sec:evaluation} 
we present the evaluation protocol followed to validate 
our method accompanied by a discussion of the results. 
Finally, we draw conclusions in Section~\ref{sec:conclusion}.

\section{Related Work}
\label{sec:relatedWork}

The method proposed in this paper has the objective of modeling 
visual compatibility between objects depicted in images by 
performing a hierarchical extraction of mid level elements. 
For this reason, we position our work w.r.t. efforts going in 
the directions of visual style modeling and visual 
representations based on mid-level elements.

\textbf{Visual Style Modeling:}
Recent work has addressed the problem of visual style modeling 
by focusing on clothing-related objects. In \cite{Bossard12}, 
a method is proposed to perform clothing description from natural 
images. To this end, face and upper body detectors are used to 
localize regions of interest. Then, an ensemble of Random 
Forests and Support Vector Machines (SVMs) are used to classify 
the clothing type and describe the clothing style, respectively.
In \cite{SimoCVPR15}, a multi-modal approach is proposed in which 
a Conditional Random Field model jointly reasons about several 
factors, e.g. outfits, user, setting, image text metadata, etc. 
to predict ``fashionability''. 
In parallel, in \cite{McAuleySIGIR15}, visual compatibility between 
clothes is modeled by learning a distance metric from CNN activations 
based on a vanilla ImageNet AlexNet between images of compatible clothes.
Just recently the method proposed in \cite{VeiKovBelMcABalBel15}, 
extends \cite{McAuleySIGIR15} by removing the logistic regression 
step and fine-tuning the entire network by integrating a Siamese 
architecture.
Different from \cite{Bossard12}, which focuses mostly on the 
description of upper-body clothing, in this work we model the 
appearance of a wider a range of clothing classes that are not exclusive 
to the upper-body, e.g. shoes, sunglasses, hats, pants, etc. In 
addition, since our method aims at discovering informative 
features for modeling visual appearance and compatibility, it does 
not require a set of attributes, e.g. colors, materials, etc., or 
classes of style to be pre-defined. 
Similar to \cite{McAuleySIGIR15,VeiKovBelMcABalBel15} our method 
uses CNN activations as a low-level feature from which a more 
informative representation is learned. Different from them, we 
obtain such representation by a hierarchical mining of mid-level 
elements. This has the advantage of providing an insight on the 
elements, or features, that drive the link between compatible 
objects. Finally, different from \cite{SimoCVPR15}, our method 
models compatibility between objects by lifting features from 
the image space exclusively. 

\textbf{Representations Based on Mid-Level Elements:}
In recent years, mid-level visual elements have been proposed
as a bottom-up means to represent visual information. These 
mid-level elements are both representative, i.e. they can be 
adapted to describe different images, and discriminative, i.e. they 
can be detected with high precision and recall. Given these 
properties, mid-level visual elements have proven to be very 
effective for several computer vision tasks such as image 
classification \cite{bossard14,Li2013,Singh2012DiscPat}, action 
recognition \cite{jainCVPR13vidrep,wangCVPR2013}, geometry 
estimation \cite{Fouhey13}, etc. More recently, mid-level visual 
elements were successfully applied to summarize large sets of 
images \cite{RematasCVPR15}. In addition, their performance was 
further improved by exploiting more advanced features at the lower 
level, combinations of activations of Convolutional Neural Networks, 
and association rule mining \cite{LiLSH15}. In this work we propose 
a hierarchical method based on mid-level elements to model visual 
compatibility between objects depicted in images. At the base-level, 
our method gets closer to existing work by extracting mid-level 
elements in a class-specific fashion. Different from existing work 
at the top-level, our method exploits co-occurrences of base-level 
elements between images of compatible objects in order to discover 
a set of patterns or rules that link such objects.

\section{Background}
\label{sec:background}

Following the original definition by Agrawal et al.~\cite{AgrawalMining93},
the problem of association rule mining is defined in the following 
terms.
Let $I=\{i_1, i_2,\ldots,i_n\}$ be a set of $n$ attributes 
called items and $D = \{t_1, t_2, \ldots, t_m\}$ be a set of 
$m$ transactions called the database with $m>>n$.
Each transaction in $D$ has a unique transaction $id$ and contains 
a subset of the items in $I$. 
A rule is defined as an implication of the form $X \Rightarrow Y$, 
where $X, Y \subseteq I$ and $X \cap Y = \emptyset$. Every rule is 
composed by two different sets of items, also known as \textit{itemsets}, 
$X$ and $Y$, where $X$ is called antecedent or left-hand-side 
(LHS) and $Y$ consequent or right-hand-side (RHS).
The goal of association rule mining is to identify strong rules between 
items in $D$ using some measures of interestingness.
In order to select interesting rules from the set of all possible 
rules, constraints on various measures of significance and interest 
are used. The best-known constraints are thresholds on support 
$supp(X \Rightarrow Y)$ and confidence $conf(X \Rightarrow Y)$. 
Formally speaking, 

\begin{equation}
    supp(X \Rightarrow Y) = supp(\{X \cup Y\}) = \frac{|\{X \cup Y\} \in D|}{m},
\end{equation}

where $|.|$ is the cardinality operator and $supp(X \Rightarrow Y) \in [0,1]$
The support $supp(X \Rightarrow Y)$ of a rule, corresponds to the proportion 
of transactions in which the itemset $\{X \cup Y\}$ occur. Similarly, the 
confidence of the rule $conf(X \Rightarrow Y)$ is defined as:

\begin{equation}
    conf(X \Rightarrow Y) = \frac{supp(\{X \cup Y\})}{supp(X)} 
\end{equation}

\section{Proposed Method}
\label{sec:proposedMethod}
\begin{figure}
	\centering
		\includegraphics[width=0.98\textwidth]{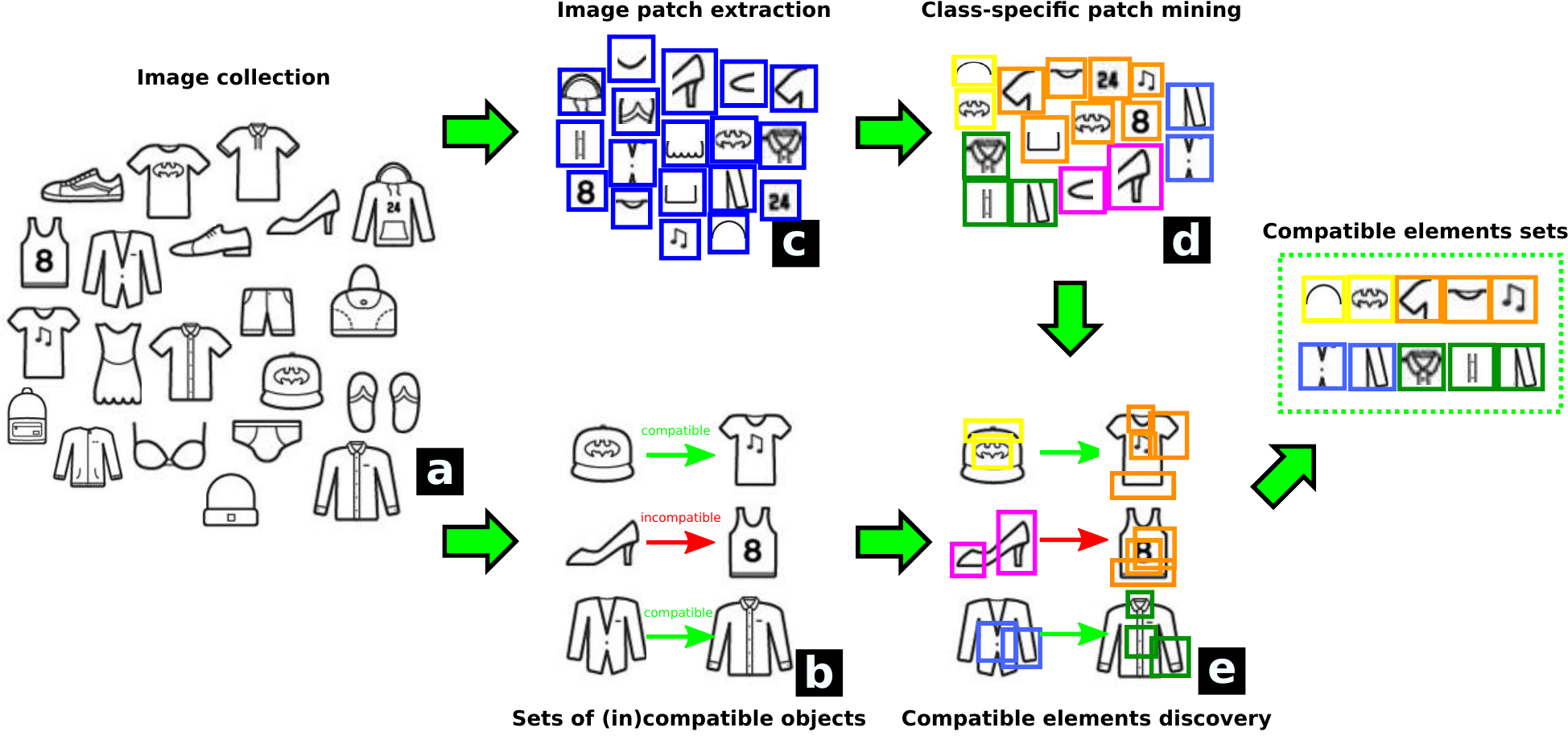}
		\caption{The proposed method consists of three steps. 
		Given a set of object images (a) accompanied with object-class
		labels, and links between them describing the (in)compatibility 
		between them (b), we proceed as follows. First, we extract 
		image patches for each image in the collection and compute CNN 
		activations from each patch (c). Second, we perform mining on 
		the CNN activations from images corresponding to each object 
		class (d), producing a set of activation patterns (base-level 
		elements) that describe each of the classes of interest. 
		Third, given a set of image pairs, the base-level elements are 
		detected in each image and sets of compatible base-level elements 
		are discovered (e) via association-rule mining~\cite{AgrawalMining93}.}
\vspace{-0.6cm}
\label{fig:methodPipeline}
\end{figure}

In this section we take a deeper look at the parts that constitute 
the proposed method to model visual compatibility. This method can 
be summarized in three steps. First, given a set images of the class 
of interest (Figure~\ref{fig:methodPipeline}.a), initial features 
are computed from image regions sampled from the images (Figure~\ref{fig:methodPipeline}.c). 
In the second step. these initial features are mined in order to 
obtain a subset of base-level elements that can be used to describe 
each object class (Figure~\ref{fig:methodPipeline}.d). 
In the third step, given a set of image pairs of (in)compatible objects
(Figure~\ref{fig:methodPipeline}.b), base-level elements with high co-occurrence 
on the image pairs are discovered (Figure~\ref{fig:methodPipeline}.e). 
These discovered pairs constitute top-level relational elements that 
can be use to describe visual compatibility between objects. 
Now we will provide a deeper presentation of the proposed method.

\subsection{Modeling Object Appearance}
\label{sec:objDescr}
Given a set of object classes $C=\{c_1,c_2,...,c_n\}$, at the base-level, 
the proposed method aims at modeling the appearance of different 
instances one class at a time. Until recent years, 
a traditional method to model visual appearance was by using low-level 
local features \cite{TuytelaarsLocalFeatures2008} such as SIFT 
\cite{loewSIFT} or HOG \cite{dalalTriggsCVPR05}. In recent years, a novel 
paradigm has emerged where mid-level visual elements are used as 
an intermediate representation between low-level features and high-level 
class-related semantics. Inspired by the success of this new paradigm, 
at the base-level, our method discovers a set of elements that can be 
used to describe different instances of a specific object class $c_i$. 
%

\subsubsection{Mining CNN Activations:}
Given an image of an object, we uniformly sample a set of regions $R=
\{r_1,r_2,\dots,r_m\}$ and for each region $r_i$ we compute CNN activations 
as an initial feature. The main motivation behind the use of CNN activations 
as initial feature is their high performance~\cite{Krizhevsky12} in challenging 
classification tasks such as the ILSVRC image classification challenge~\cite{RussakovskyILSVRC15}.
Furthermore, in recent years there is clear evidence that these features, 
initially trained for classification, can be fine-tuned to be used for
object detection \cite{girshick14CVPR,kaiming14ECCV} and instance retrieval 
\cite{Sharif14}. This makes CNN features a strong means to describe appearance.

\begin{figure}
	\centering
		\includegraphics[width=0.6\textwidth]{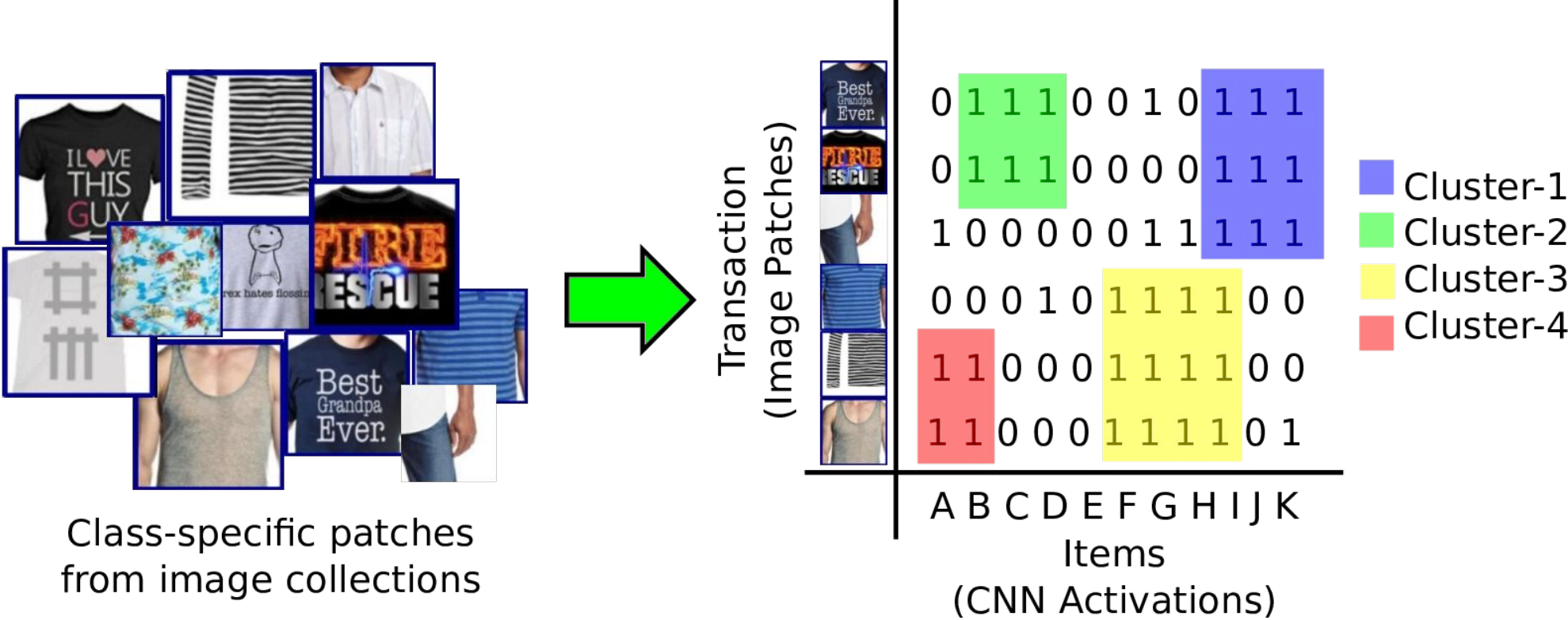}
		\caption{Transaction-Item Matrix construction for mining patterns of 
		CNN activations that describe the object classes of interest.}
\label{fig:objectAppearanceMining}
\end{figure}

Given a set of image regions $R$ sampled from the image collection,  
with their corresponding CNN activations $F$, we mine CNN activation patterns $P'$
in a class-specific fashion. To this end, given a set of regions $R^c$, 
from a set of images depicting objects of class $c$, we define the transaction-item 
matrix where the transactions correspond to the image regions $r^c_i \in R^c$ 
and the items are defined by each of the dimension indices from the CNN activations 
(see Figure~\ref{fig:objectAppearanceMining}). Following its construction, this 
matrix is further binarized by taking the subset consisting of the top-$k$ activations per transaction. As presented in \cite{LiLSH15,DosovitskiyB15,Agrawal2014}, no relevant information is lost during the binarization step, since discriminative information from the CNN activations is mostly embedded in the dimension indices of the largest magnitudes. We recommend reading \cite{LiLSH15} for a deeper analysis on this aspect.
A set of class-specific CNN activation patterns $P'^c$ is obtained as an outcome of this mining procedure.

\subsubsection{Retrieving Base-level Visual Elements:}
\begin{figure}[h!]
	\centering
		\includegraphics[width=0.86\textwidth]{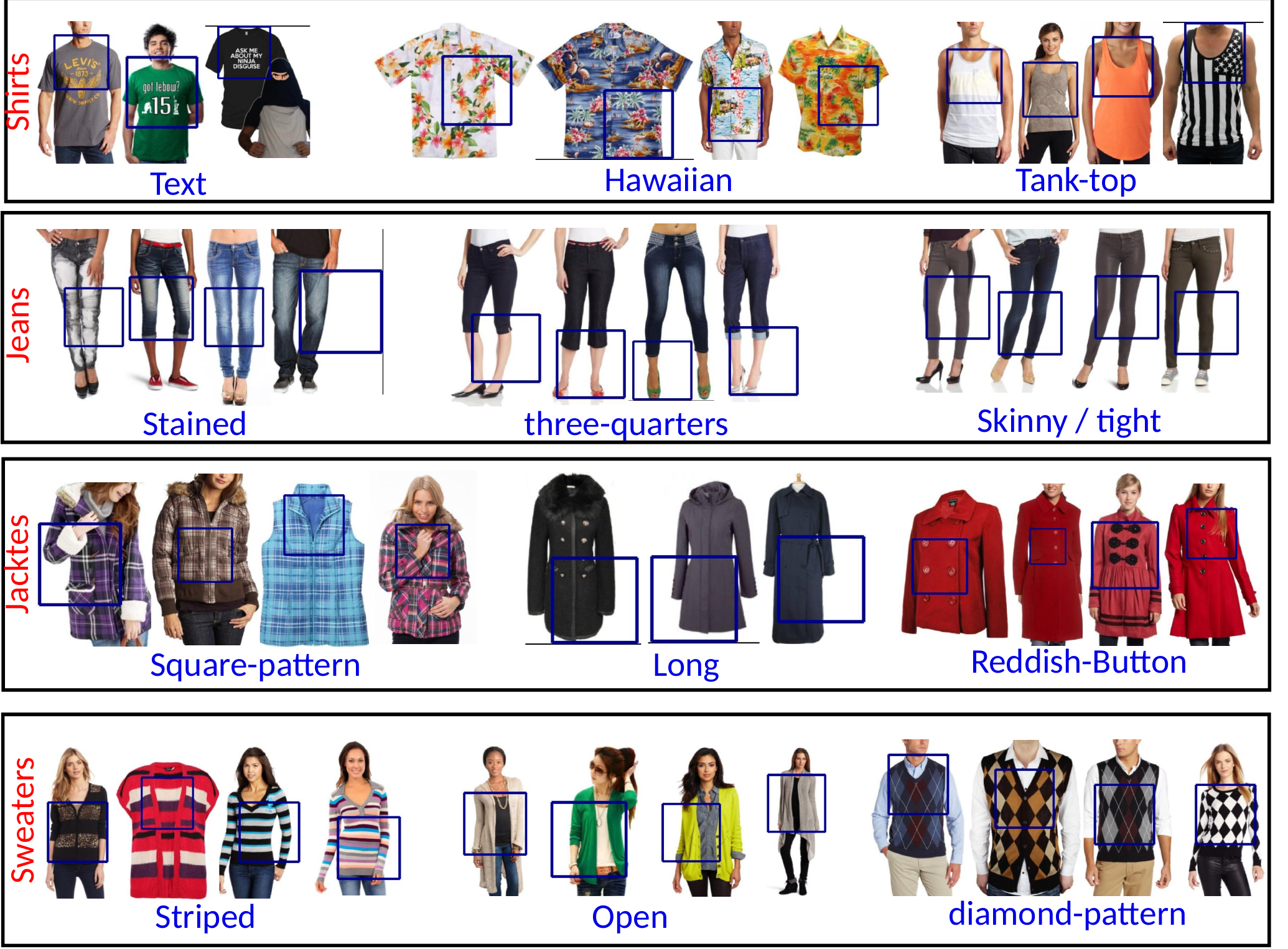}
		\caption{Some of the clusters defined by the base-level visual elements for the \textit{shirts, jeans, coats \& jackets}, and \textit{sweaters} product classes. For each of the clusters we have added 
		a text caption (in blue) of the style that they seem to encode. Furthermore, each image within each cluster is marked with the patch that encodes such style.}
\label{fig:productStyleClusters}
\end{figure}

Starting from the set of activation patterns $P'^c$, discovered in the previous step, we perform the following two steps. First, we remove redundant patterns that fire on the same set of items (CNN activation indices), thus, producing a reduced set of patterns $P'^c$. Then, the next step consists of extracting a set of visual elements $V'^c \in P'^c$. The visual elements $v'^c_i \in V'^c$ cover all the image patches that contain the pattern $p'^c_i \in P'^c$. This can be retrieved effectively via an inverted file index. 
As a result, we obtain a set of base-level visual elements $V'^c$ which describe different visual characteristics of instances of class $c$. Please see Figure~\ref{fig:productStyleClusters} for some groups of images clustered by their base-level elements. Notice how some of these elements effectively describe ``style'' features of the classes of interest. Finally, for each base-level element $v'^c_i$ we train a LDA classifier $\theta'^c_i$ using all the image patches covered by $v'^c_i$, producing a set of LDA classifiers $\Theta'^c = \{\theta'^c_1,\theta'^c_2,\dots,\theta'^c_n\}$ for object class $c$.

\subsection{Modeling Visual Compatibility}
\label{sec:modelVisComp}

In a nutshell, our method to model visual compatibility consists of 
discovering patterns of co-occurring activations of the base-level elements 
extracted in Section~\ref{sec:objDescr}, and considers the level to which 
they occur in a pair of test images in order to measure their compatibility.

\subsubsection{Base-level Data Encoding:}
Given a set of image pairs $(I_{(a)},I_{(b)})$ accompanied with binary compatibility labels $l \in \{0,1\}$, for each image, we sample a set of regions $R$ and compute CNN activations following a similar procedure as stated in Section~\ref{sec:objDescr}. Then, each image from the pair is re-encoded using the LDA classifiers $\Theta'^c$ corresponding to the class $c$ of the object of interest. To this end, we compute the response $\theta'^c_i(r)$ of each classifier $\theta'^c_i \in \Theta'^c$ for each image region $r \in R$ and store the maximum responses and the corresponding ids of the regions that produced them. 
This produces a new representation defined over base-level elements $V'^c$ with dimensionality equal to the number of classifiers $\theta'^c_i(r)$ for each image. 

\subsubsection{Mining Base-level Activations:}
\begin{figure}
	\centering
		\includegraphics[width=0.6\textwidth]{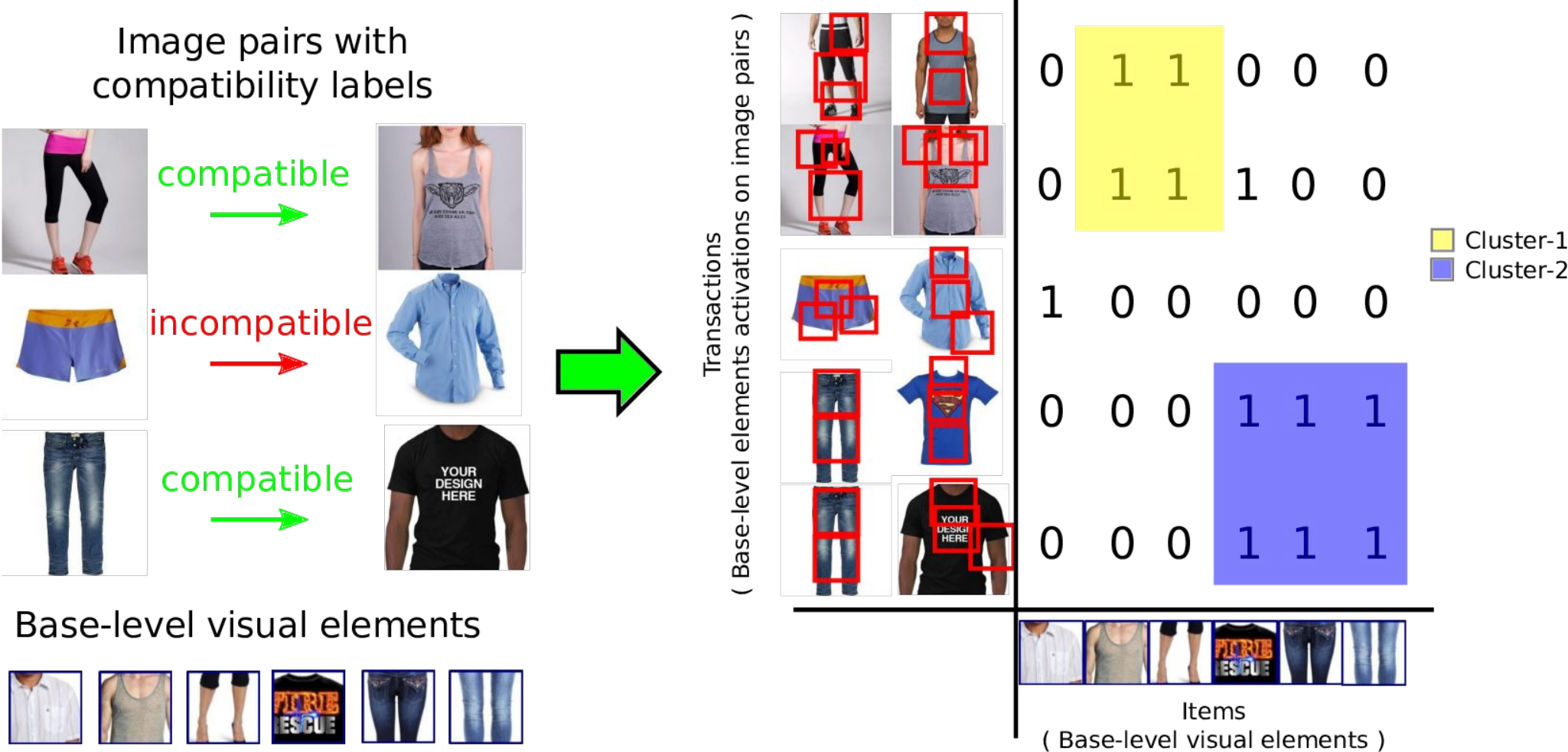}
		\caption{Transaction-Item Matrix construction for mining patterns of co-occurring base-level element activations that describe links between objects of specific classes.}
\label{fig:compatibilityMining}
\end{figure}

In order to discover co-occurring activations of the base-level elements, we perform association-rule mining \cite{AgrawalMining93} by considering the base-level representation of the image pairs and their respective compatibility labels $l$.
At this part of the algorithm, we construct a new the transaction-item matrix (see Figure~\ref{fig:compatibilityMining}) where each transaction is defined by the concatenated base-level representations from the image pairs. Each transaction is further extended with two additional bins which are used to indicate whether the objects are compatible or not. Moreover, the image pairs defining the transactions cover both compatible and incompatible pairs. The items of this matrix are defined by the set of base-level elements defined in the previous section. Following the association-rule mining \cite{AgrawalMining93} approach we mine rules of the form 
$[ \Theta'^{c_a}(R_a), \Theta'^{c_b}(R_b) ] \Rightarrow (l_{ab}=1)$. The antecedent of these rules are the concatenated responses of the base-level elements from object images $(I_a,I_b)$ of class $c_a$ and $c_b$, respectively. The consequent $l_{ab}$ is the binary compatibility label of object pair $(I_a,I_b)$, which we require to be positive $(l_{ab}=1)$ .

This mining process is performed, independently, for all the combinations of object classes $(c_i,c_j) \in C$. Hence, the transactions of the transaction-item matrix only considers image pairs where images from both classes are involved. Furthermore, the items from such matrix only cover base-level elements $\{V'^{c_i},V'^{c_j}\}$.  
Finally, this matrix is binarized by selecting the top-$k$ activations within the base-level representation of each object on each pair.
For all class combinations $(c_i,c_j) \in C$, a set of co-occurring activation patterns $P''^{(c_i,c_j)}$ covering examples of classes $(c_i,c_j)$ is obtained as a result of this mining step.

\subsubsection{Retrieving Top-level Visual Elements:}
Starting from the set of activation patterns $P''^{(c_i,c_j)}$, discovered in the previous step, the next step consists of extracting a reduced set of top-level visual elements $V''^{(c_i,c_j)} \in P''^{(c_i,c_j)}$. These top-level elements are retrieved in a similar fashion as for the base-level elements $V'^c$ (Section~\ref{sec:objDescr}), i.e. via an inverted file index.
As a result, we obtain a set of top-level visual elements $V''^{(c_i,c_j)}$ which describe links between base-level visual elements $V^c$ for compatible objects. Hence, their occurrence on a given pair of images can be used to indicate whether the objects depicted on such images are compatible or not.
Finally, for each top-level element $V''^{(c_i,c_j)}$ we train a LDA classifier $\theta''^{(c_i,c_j)}$ using all the image pairs covered by $v''^{(c_i,c_j)}$, producing a set of classifiers $\Theta''^{(c_i,c_j)} = \{\theta''^{(c_i,c_j)}_1,\theta''^{(c_i,c_j)}_2,\dots,\theta''^{(c_i,c_j)}_n\}$ for object pairs belonging to classes $(c_i,c_j)$. Each of these classifiers is used to estimate the level to which a specific top-level element occurs on a pair of images.

\subsection{Inference}
\label{sec:inference}

Having the sets of base-level $\Theta'^c$ and top-level $\Theta''^{(c_i,c_j)}$ classifiers, inference is a straight forward process (Algorithm~\ref{alg:inference}).
Given a pair of test images $(I_a,I_b)$ depicting object of classes $(c_a,c_b)$, respectively. As a first step we uniformly sample regions $r_a$ and $r_b$ from each image producing the respective sets $R_a$ and $R_b$. Then, we compute CNN activations $f_a = f(r_a)$ and $f_b=f(r_b)$ for each for each region in the sets $R_a$ and $R_b$.  
Then using the base-level classifiers  $(\Theta'^{c_a},\Theta'^{c_b})$, 
for classes $c_a$ and $c_b$, we compute the encoding $\Gamma_{ab}$ for the image pair $(I_a,I_b)$  by taking the maximum response for each classifier over the different CNN activations in $F_a$ and $F_b$ corresponding to the image regions $r_a \in R_a$ and $r_b \in R_b$, respectively. Hence $ \Gamma_{ab} = [max(\Theta'^{c_a}(F_a)),max(\Theta'^{c_b}(F_b))]$. Finally, using the top-level LDA classifier $\Theta''^{(c_a,c_b)}$, we compute the compatibility score between $I_a$ and $I_b$ by taking the maximum response from the set of responses defined by $\Theta''^{(c_a,c_b)}(~ \Gamma_{ab} ~) = [\theta''^{(c_a,c_b)}_1(\Gamma_{ab}),\theta''^{(c_a,c_b)}_2(\Gamma_{ab}),\dots,\theta''^{(c_a,c_b)}_n(\Gamma_{ab})]$.

\begin{algorithm}
\caption{\label{alg:inference} Visual Compatibility Estimation}
\textbf{\footnotesize Given}{\footnotesize \par}

\begin{itemize}
\vspace{-0.25cm}
\item {\footnotesize A pair of images $(I_a,I_b)$ of classes $(c_a,c_b)$, respectively.}{\footnotesize \par}
\end{itemize}

\vspace{-0.25cm}
\textbf{\footnotesize Steps}{\footnotesize \par}

\begin{enumerate}
\item {\footnotesize \textbf{Region Sampling:}
uniformly sample regions $R_a$ and $R_b$ on each image.}{\footnotesize \par}

\item {\footnotesize \textbf{Initial Feature Encoding:} 
compute CNN activations $f_a \in F_a$ and $f_b \in F_b$ for each region $r_a \in R_a$ and $r_b \in R_b$.}{\footnotesize \par}

\item {\footnotesize \textbf{Base-level Scoring:} 
compute the responses $\theta'^{c_a}(f_a)$ and $\theta'^{c_b}(f_b)$ 
using the base-level LDA classifiers $(\Theta'^{c_a},\Theta'^{c_b})$ 
for classes $c_a$ and $c_b$, respectively.}{\footnotesize \par} 

\item {\footnotesize \textbf{Base-level Encoding:} 
concatenate the maximum response for each classifier $(\Theta'^{c_a},\Theta'^{c_b})$ producing the joint response $ \Gamma_{ab} = [max(\Theta'^{c_a}(F_a)),max(\Theta'^{c_b}(F_b))]$.}{\footnotesize \par}

\item {\footnotesize \textbf{Top-level scoring:} using the top-level LDA classifier $\Theta''^{(c_a,c_b)}$ compute the compatibility score between $I_a$ and $I_b$ by taking the maximum response from the set of responses $\Theta''^{(c_a,c_b)}(~ \Gamma_{ab} ~)$.}{\footnotesize \par}			  			  
\end{enumerate}

\end{algorithm}

\section{Implementation Details}
\label{sec:implementationDetails}

\textbf{Modeling Object Appearance:}
When modeling object appearance (Section~\ref{sec:objDescr}), each of the images is resized proportionally so that their smaller dimension is 256 pixels. Then, image patches are extracted by using a sliding window approach with window size 128x128 pixels and a stride of 32 pixels. On average, a set of 40 windows are extracted per image. We compute CNN activations using \textit{Caffe}~\cite{jia2014caffe} in combination with the \textit{CaffeRef} model~\cite{jia2014caffe}.
During mining, we binarize the CNN activations of each transaction by selecting the top 20 activations ($k=20$).  We look for patterns $P^c$ composed of 3 to 6 items, i.e. $3 \leq length(p^c) \leq 6, \forall p^c \in P^c$. Finally, we aim to extract at most 4000 base-level visual elements $v^c_i$ per class.

\textbf{Modeling Visual Compatibility:}
When modeling visual compatibility (Section~\ref{sec:modelVisComp}), the transaction-item matrix is binarized by selecting the top 10 activations ($k=10$) of the base-level representation of each object prior to their concatenation to define a transaction.
During mining, we search for patterns composed of 3 to 6 items, i.e. $3 \leq length(p''^{(c_i,c_j)}) \leq 6$. Moreover, we enforce that the items defining each of the patterns $p''^{(c_i,c_j)}$ cover base-level visual elements from both classes. At this level, we perform association-rule mining with support $supp(p''^{(c_i,c_j)}) \geq 0.05\%$ and  confidence $conf(p''^{(c_i,c_j)}) \geq 75\%$. Finally, similar to the case of base-level elements, we aim at extracting at most 4000 top-level visual elements in total.

\section{Evaluation}
\label{sec:evaluation}
We conduct experiments on the dataset\footnote{http://jmcauley.ucsd.edu/data/amazon/} collected in \cite{McAuleyKDD15} which is based on the Amazon web store. 
This dataset was collected between May 1996 and July 2014, 
thus producing a large scale set of 9.4 million product images.
Each product is accompanied by an image and metadata covering 
product ID, sales-rank, brand, and co-purchasing product links.
These links, e.g. ``also bought'',``bought together'',
``also viewed'', are defined by users of the web store while 
interacting with such products.
In our experiments we focus on the subset of this dataset 
defined in \cite{VeiKovBelMcABalBel15} which covers around 1 million 
products belonging to the parent class ``Clothing, Shoes and Jewelry''.
This subset includes a set of $\sim$2.5 million product 
pairs with compatibility labels defined from product co-purchase 
links covering the relations ``bought together'' and ``also bought''.  
These compatibility labels are binary labels stating whether two 
specific products are compatible or not. 
Since the proposed method is focused on modeling visual 
compatibility, we limit ourselves to only consider the image and 
the class label for each product and ignore any additional metadata.

We focus our experiments considering the classes \textit{accessories, jeans, 
pants, shirts, shoes, shorts, skirts, and tops \& tees} covering $\sim$500K 
images. Total training time took below 12 hours. 
Following the evaluation protocol from \cite{VeiKovBelMcABalBel15}, 
we use as performance metric the area under the curve (AUC) defined 
by the False Positive Rate (FPR) vs. the True Positive Rate (TPR) 
on the compatibility estimation task.
We report qualitative results in Figures~\ref{fig:qualitativeResultsExtended},  \ref{fig:qualitativeResults} and \ref{fig:objectContextModeling} .

\subsection{Quantitative Results}
In this section we analyze the quantitative performance of the proposed
method and compare w.r.t. the method proposed in \cite{VeiKovBelMcABalBel15}. 
For this purpose, given the testing set of image pairs, we run the method 
from \cite{VeiKovBelMcABalBel15} using the publicly available code\footnote{http://vision.cornell.edu/se3/projects/clothing-style/}.
The method from \cite{VeiKovBelMcABalBel15} has clearly superior 
performance (0.804 \small{AUC}) than the one proposed in this paper (0.655 \small{AUC}).
However, despite its high performance, since the method from \cite{VeiKovBelMcABalBel15} operates at the image level it has a reduced 
output in terms of ``explaining'' the reasons that make the compared 
objects compatible. On the contrary, while still being able 
to estimate the compatibility up to some level, our method has the 
advantage of providing some insights related to the object characteristics 
that drive such compatibility (see Section~\ref{sec:qualitativeResults}).

\subsection{Qualitative Results}
\label{sec:qualitativeResults}
\begin{figure}[h!]
	\centering
		\includegraphics[width=0.9\textwidth]{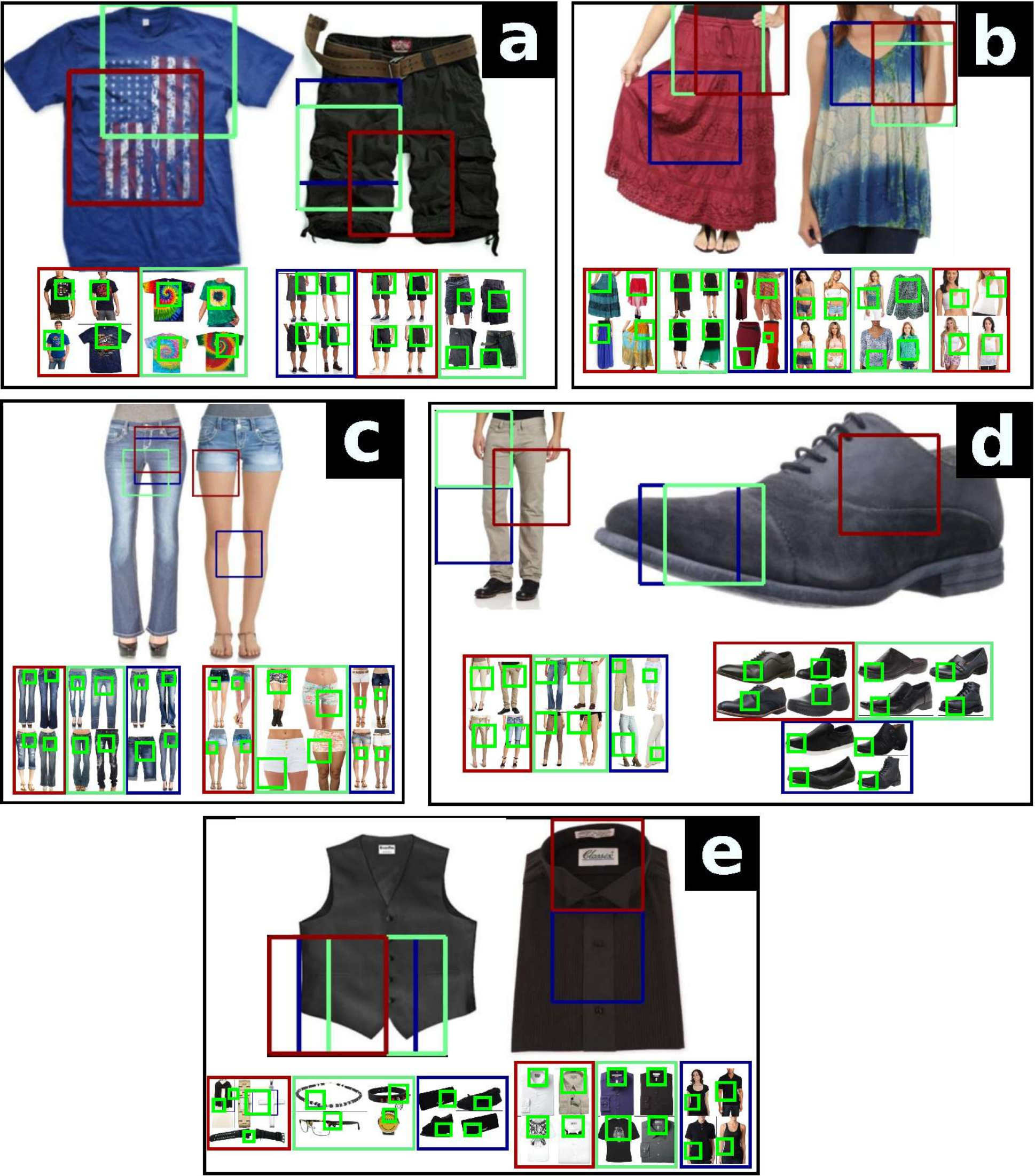}
		\caption{Some examples of accurately predicted compatible objects.
			 For each object the regions of the top-3 base-level elements
			 are indicated with their scores color-coded in jet scale.
			 In addition, for each base-level element, we present a subset of random examples that compose it.
			 Note how some base-level elements effectively describe some of the features that define the link between the objects (a-b), while for others (object\#1 on (e)) this information is not so clear.}
\label{fig:qualitativeResultsExtended}
\end{figure}

\begin{figure}[h!]
	\centering
		\includegraphics[width=0.9\textwidth]{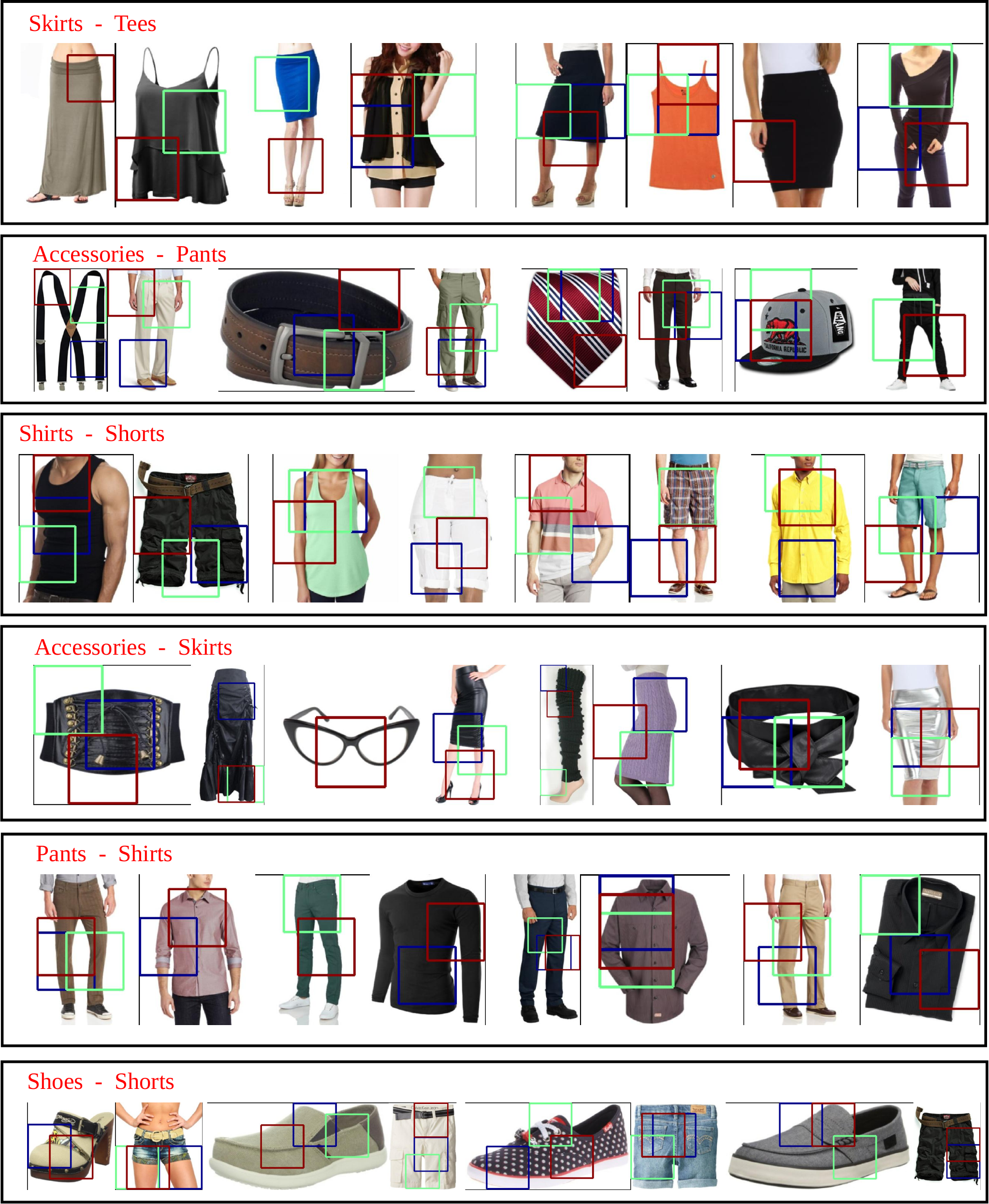}
		\caption{Some examples of accurately predicted compatible objects.
			 For clarity, for each object the regions of only the top-3 base-level elements are indicated with their scores color-coded in jet scale. Please refer to the supplementary material for an extended list of examples.}
\label{fig:qualitativeResults}
\end{figure}

In Figure~\ref{fig:qualitativeResultsExtended} we present some examples where 
object compatibility has been predicted accurately. For each example we draw 
the regions representing the top-3-scoring base-level elements corresponding to 
each image. Moreover, we extend each example by showing a random subset of 
training image regions that define such base-level elements. Notice how for 
some of the examples, some of the base-level elements provide details regarding the 
features that characterize each object and drive the compatibility. For example, 
Dark colorful logo \textit{shirt} with dark loose pocketed \textit{shorts} (Figure~\ref{fig:qualitativeResultsExtended}.a), reddish loose \textit{skirt} with blue/greenish \textit{tee} without sleeves (Figure~\ref{fig:qualitativeResultsExtended}.b).
In addition, in Figure~\ref{fig:qualitativeResults} we show a larger set of 
qualitative examples. For clarity, for each of the images of the figures, 
we show a reduced set base-level elements which have the highest response 
when measuring compatibility between the objects.
Please refer to the supplementary material for an extended list of examples

\subsection{Encoding Context Cues}
\label{sec:objectContextModeling}
\begin{figure}[h!]
	\centering
		\includegraphics[width=0.9\textwidth]{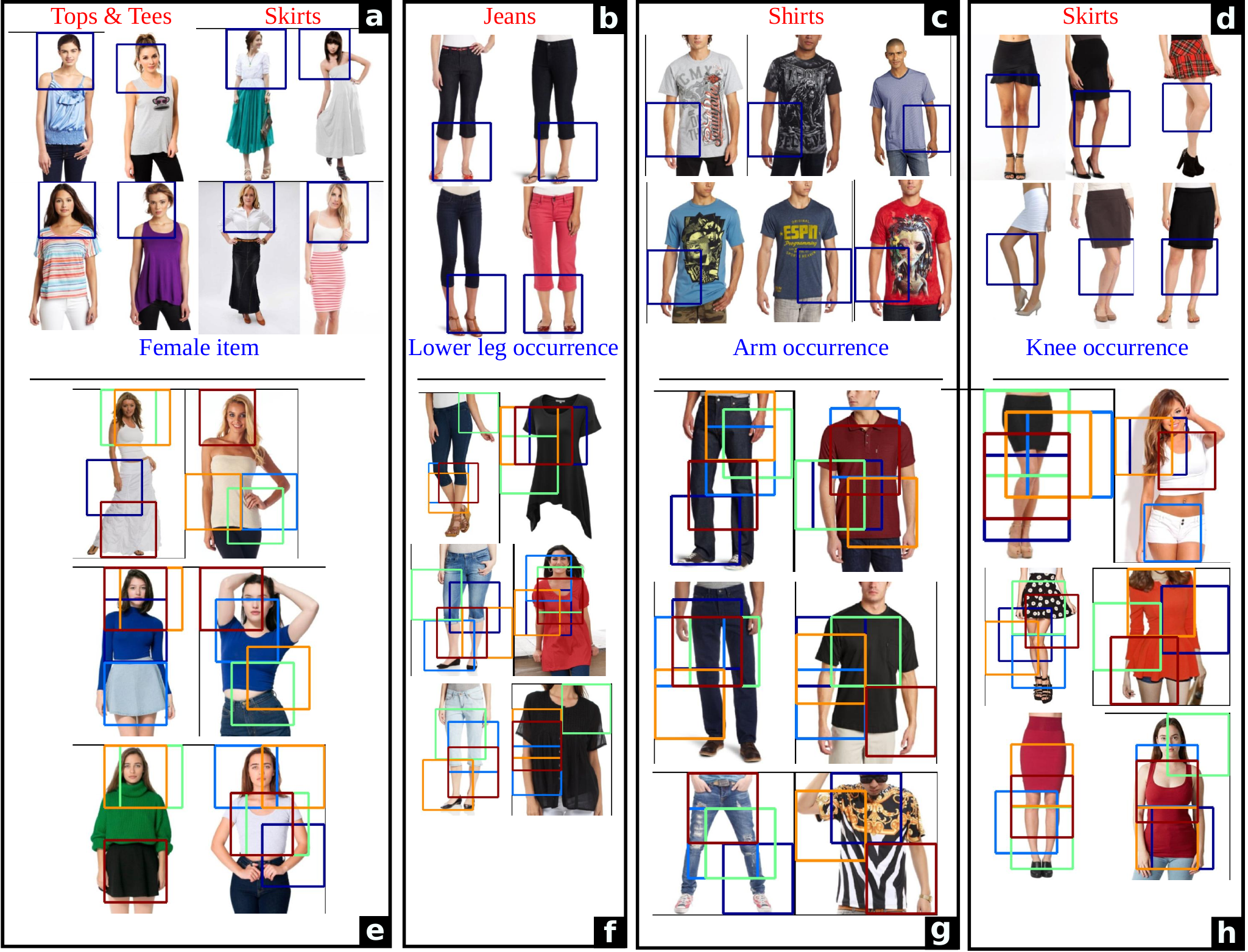}
		\caption{Examples of mid-level elements encoding contextual information at both the base~(a-d) and top~(e-h) levels, respectively.
		The top row shows the region (blue) of the base-level element. The bottom row shows comparisons between images with the top scoring base-level elements that describe the compatibility color-coded in jet scale.
		Note how the method is able to encode that \textit{skirts} and some \textit{tops} are female products (a) , and that wearing 3/4 \textit{jeans}, \textit{shirts} and mini \textit{skirts}, leave the feet, arms and knees, exposed, respectively (b-d) . Moreover, detecting a female face can produce a strong link when comparing 
		\textit{skirts} and some \textit{tops} products (e). Similarly, other  base-level elements that encode context information can serve as a strong indicator of compatibility (f-h). 
		}
\label{fig:objectContextModeling}
\end{figure}

A deeper visual inspection to the mid-level elements, at both the base 
and top levels, reveals that the proposed method is taking into account 
information from ``external'' regions that do not directly overlap with 
the object of interest. As Figure~\ref{fig:objectContextModeling} shows, 
these external regions are used for different purposes. That is, at the 
base-level, they are used to describe extrinsic features from the products,
such as that some types of \textit{tops \& tees} as well as \textit{skirts} 
are products exclusively for women (Figure~\ref{fig:objectContextModeling}.a). Likewise, wearing a three-quarter \textit{jeans} reveals the feet of the user, 
while \textit{shirts} and some type of (mini) \textit{skirts} reveal the arms 
and the knees, respectively (Figure~\ref{fig:objectContextModeling}.b-d). 
Furthermore this information seems to play an important role when linking products 
via compatibility relations (see Figure~\ref{fig:objectContextModeling}.e-h).
For example, in Figure~\ref{fig:objectContextModeling}.e, the occurrence of 
a female face produces a high response when measuring the compatibility 
between \textit{skirt} and \textit{tops \& tees} products.
This strongly suggests that the proposed method is able to lift contextual 
information from the images of the products. In the current setting, the Amazon 
dataset, almost all the images have an uniform-color background. Thus, the only 
source of contextual cues are the persons that are wearing the clothes in some 
images. Given the fact that clothing style is driven by several factors (e.g. 
shorts, tops, sandals, and t-shirts are usually worn on summer-like locations 
like beaches, suits and long dresses are commonly worn indoors, etc.) the 
capability of lifting contextual information embedded in the images, is a 
feature with promising potential when operating on images in the wild.

\subsection{Discussion}

In this paper we presented a hierarchical method to discover mid-level 
visual elements with the objective of modeling visual compatibility 
between objects. The main features of the proposed method can be divided 
into two groups. On the one hand, the proposed method extracts a meaningful 
representation that not only describes different variations, or style, 
in which the objects of interest may occur (see Figure~\ref{fig:productStyleClusters}), but it also reveals the features 
from the objects that define the compatibility between them. On the other hand, 
there is some evidence (Section~\ref{sec:objectContextModeling}) that the proposed method is able to take advantage of contextual information embedded in the images when modeling both object appearance and compatibility.

In addition, since the proposed method follows a class-driven processing approach, 
the whole pipeline can be divided into sub-processes focusing on specific object classes. As a result, each sub-process operates on a smaller volume of data which results in faster processing times. In the event that new object classes are needed to be taken into account, a total of $|C|+1$ processes should be launched. From this total, one process focuses on modeling the appearance of the new class, while the other $|C|$ (number of existing classes) processes focus on modeling the compatibility of the new class w.r.t. the existing classes in the system.  Moreover, there are 
some object classes which are never linked, e.g. swimming pants \& tuxedos, watches \& underwear, etc., for which modeling compatibility is not required. This sets our method apart w.r.t class agnostic methods which require the whole process, which operates over the whole volume of data, to be repeated each time new classes need to be considered.

Despite the notable descriptive strength of the proposed method, there are several areas in which the proposed method can be improved. First, as mentioned in Section~\ref{sec:implementationDetails}, during the region sampling step at the base-level processing, an average of 40 fixed-size regions are sampled per image. Performing a denser sampling at different scales will improve the alignment of base-level elements. Moreover, the method would be able to inspect low-resolution object characteristics which are currently un-reachable. This should produce more informative base-level elements which benefit compatibility modeling at the top-level. Furthermore, there is clear evidence \cite{kaiming14ECCV} that this dense sampling and CNN feature computations can be performed efficiently. Second, in its current state, the method computes the compatibility score between images by taking the maximum response from the LDA classifiers based on the top-level elements (Section~\ref{sec:inference}). This removes the 
possibility of different top-level classifiers operating on a coordinated fashion, which is a likely possibility given the fact that top-level classifiers are derived from base-level elements which encode different aspects (e.g. color, texture, shape, context, etc.) of the objects being compared. Moreover. these last top-level classifiers, i.e. the ones used for making the final decision about compatibility scoring, are trained by only considering positive examples, thus having reduced discrimination against negative examples. For this reason, we proposed first to train each top-level classifier in discriminative approach in which examples of incompatible objects are also considered. Furthermore, in order to take advantage of the descriptive strength of the top-level classifiers, we propose to compute the compatibility score as a combination of their responses.

\section{Conclusion}
\label{sec:conclusion}
In this paper we presented a hierarchical method to discover 
mid-level visual elements with the objective of modeling visual 
compatibility between objects. Although, on the quantitative side,  
the proposed method achieves subpar performance, on the qualitative 
side, the method shows strong descriptive capabilities.
In this regard,the proposed method not only effectively models 
different styles in which the objects of interest may occur, but 
it also provides an insight on the characteristics of the objects 
that define compatibility between them. 
Moreover, there is evidence that suggests that the proposed method 
is able to encode and effectively exploit contextual information 
when modeling object classes and the compatibility between them.

\section{Acknowledgements}
This work is supported by the SBO Program IWT project 
``Personalised AdveRtisements buIlt from web Sources (PARIS)'' 
(IWT-SBO-Nr. 110067), and a NVIDIA Academic Hardware Grant.


\bibliographystyle{splncs}
\bibliography{egbib}

\clearpage

\title{Modeling Visual Compatibility through Hierarchical Mid-level Elements\\(Supplementary Material)} 

\author{Jose Oramas M., Tinne Tuytelaars}

\institute{KU Leuven, ESAT-PSI, iMinds}

\titlerunning{Modeling Visual Compatibility via Mid-level Elements}
\authorrunning{Oramas \& Tuytelaars}


\section{Supplementary Material}

\begin{figure}[h!]
	\centering
		\includegraphics[width=0.99\textwidth]{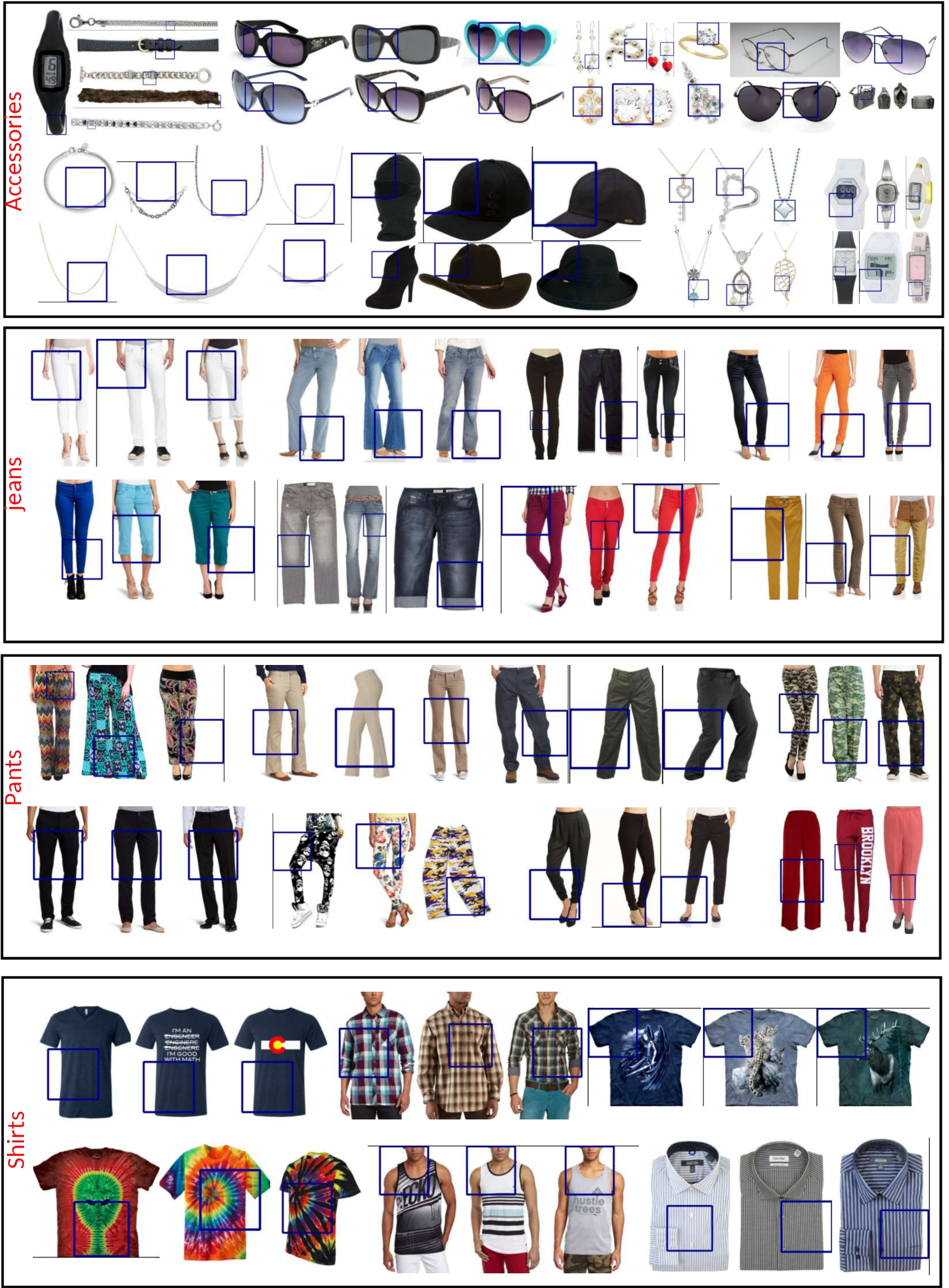}
		\caption{Some of the clusters defined by the base-level visual elements for the \textit{shirts, jeans, coats \& jackets}, and \textit{sweaters} product classes. Note how some of the clusters seem to encode some variations or 'style' in which the objects of interest may occur. Furthermore, each image within each cluster is marked with the patch that represents the base-level element.}
\label{fig:productStyleClustersP1}
\end{figure}

\begin{figure}[h!]
	\centering
		\includegraphics[width=0.99\textwidth]{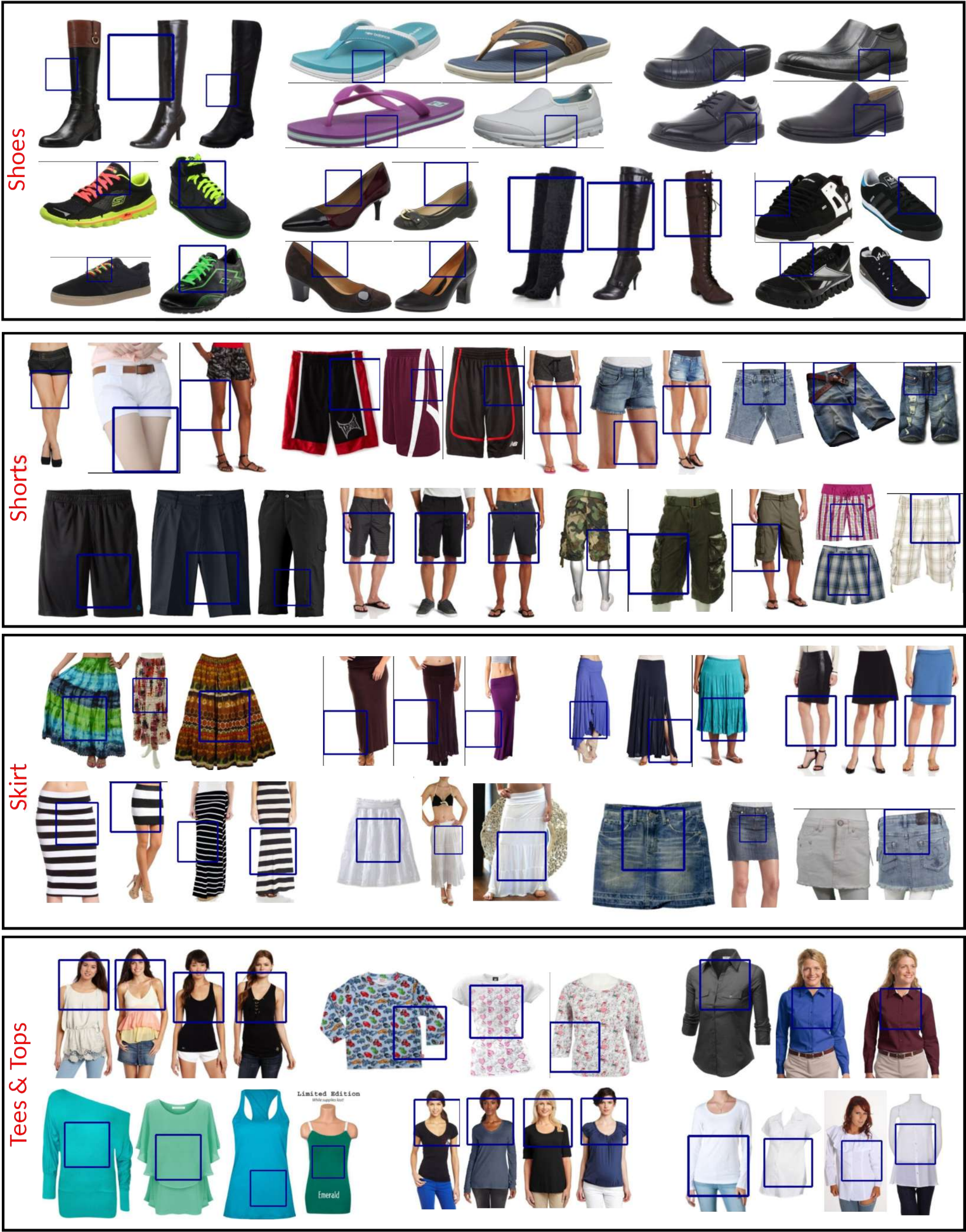}
		\caption{Some of the clusters defined by the base-level visual elements for the \textit{shirts, jeans, coats \& jackets}, and \textit{sweaters} product classes. Note how some of the clusters seem to encode some variations or 'style' in which the objects of interest may occur. Furthermore, each image within each cluster is marked with the patch that represents the base-level element.}
\label{fig:productStyleClustersP2}
\end{figure}

\begin{figure}[h!]
	\centering
		\includegraphics[width=0.99\textwidth]{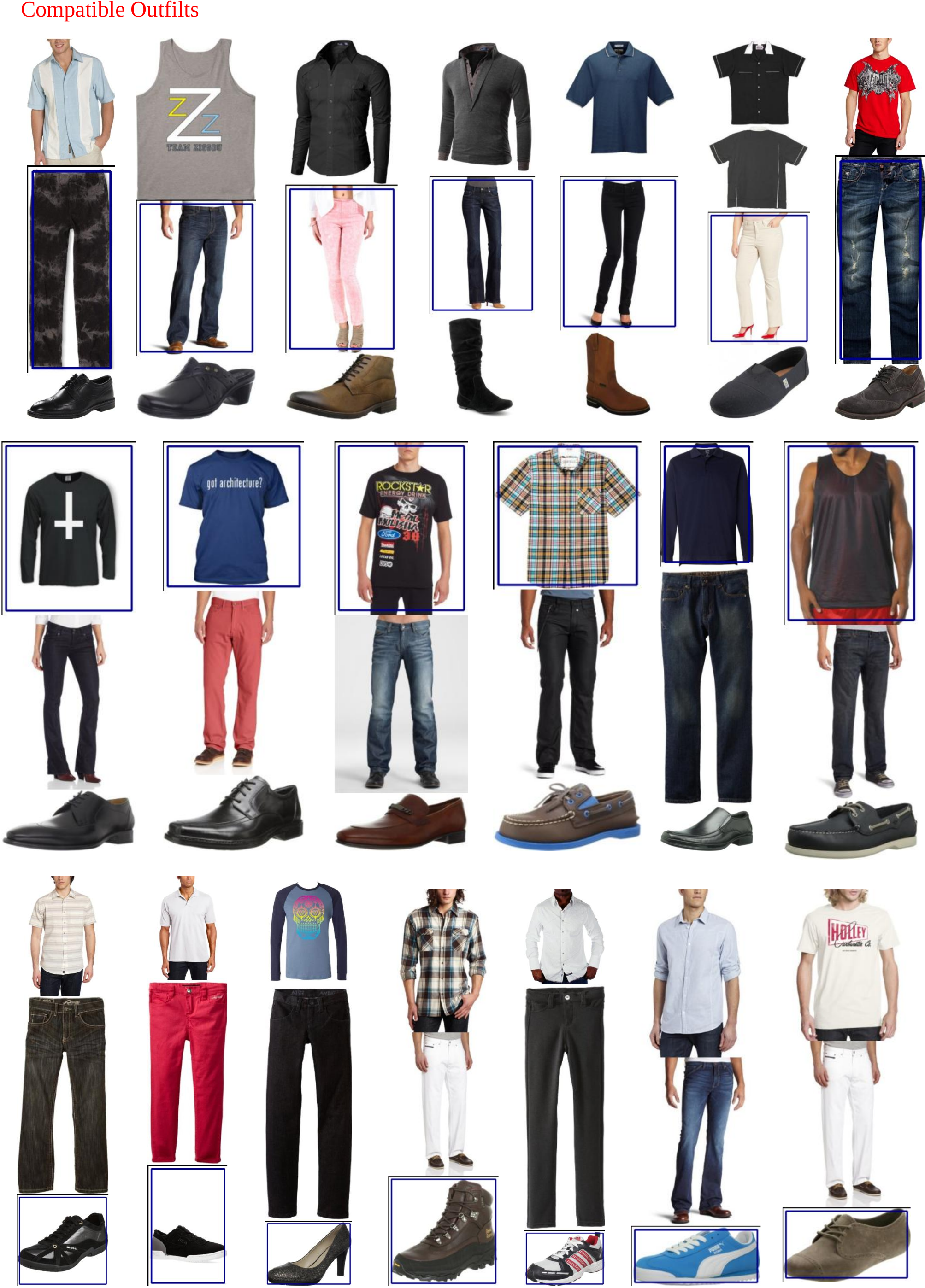}
		\caption{Outfits composed of three products from different classes predicted 
		with high compatibility based on the proposed method. Based on the query 
		product (marked in blue) compatible products of other classes are selected.}
\label{fig:generatedOutfitP1}
\end{figure}

\begin{figure}[h!]
	\centering
		\includegraphics[width=0.99\textwidth]{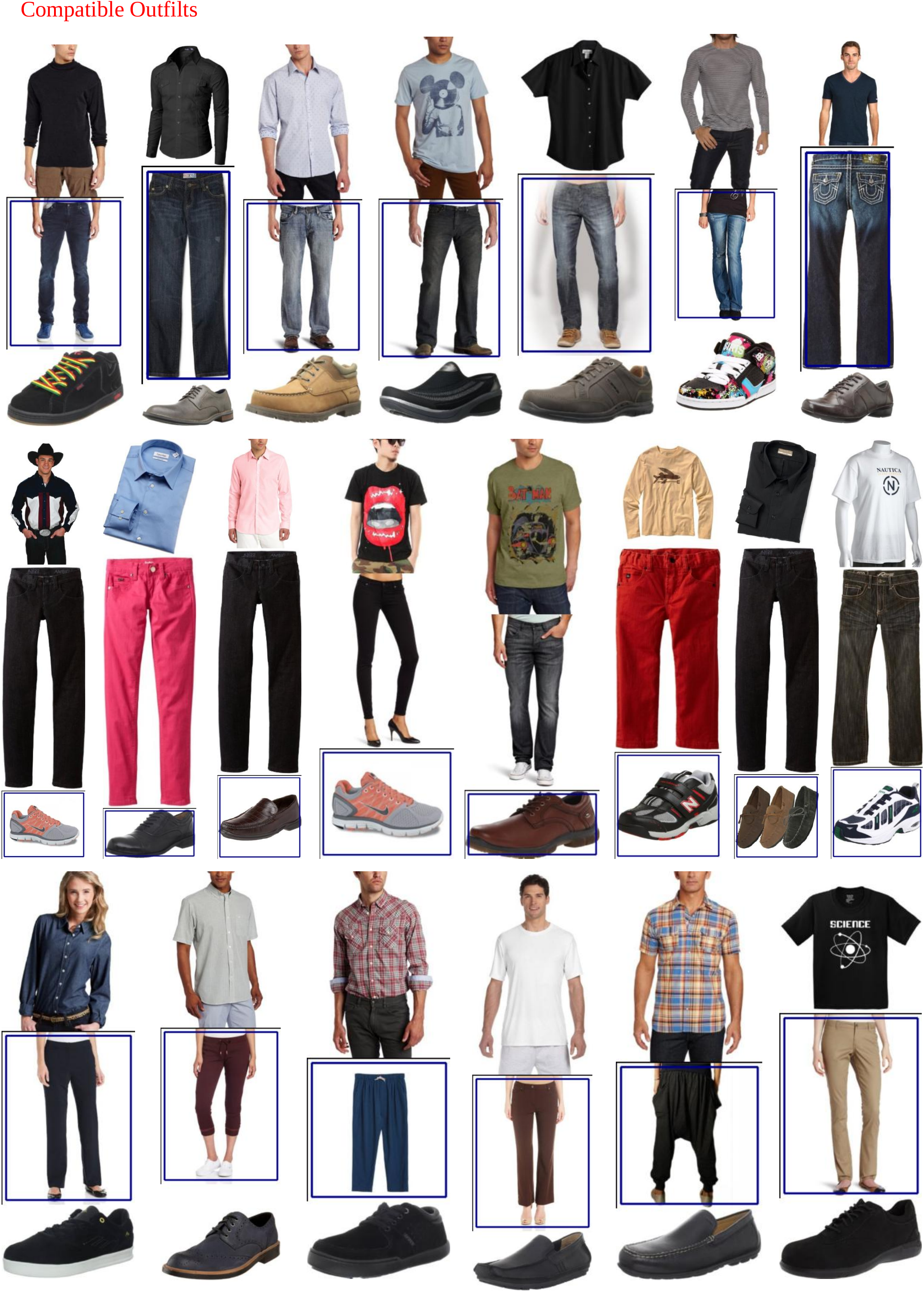}
		\caption{Outfits composed of three products from different classes predicted 
		with high compatibility based on the proposed method. Based on the query 
		product (marked in blue) compatible products of other classes are selected.}
\label{fig:generatedOutfitP2}
\end{figure}

\begin{figure}[h!]
	\centering
		\includegraphics[width=0.99\textwidth]{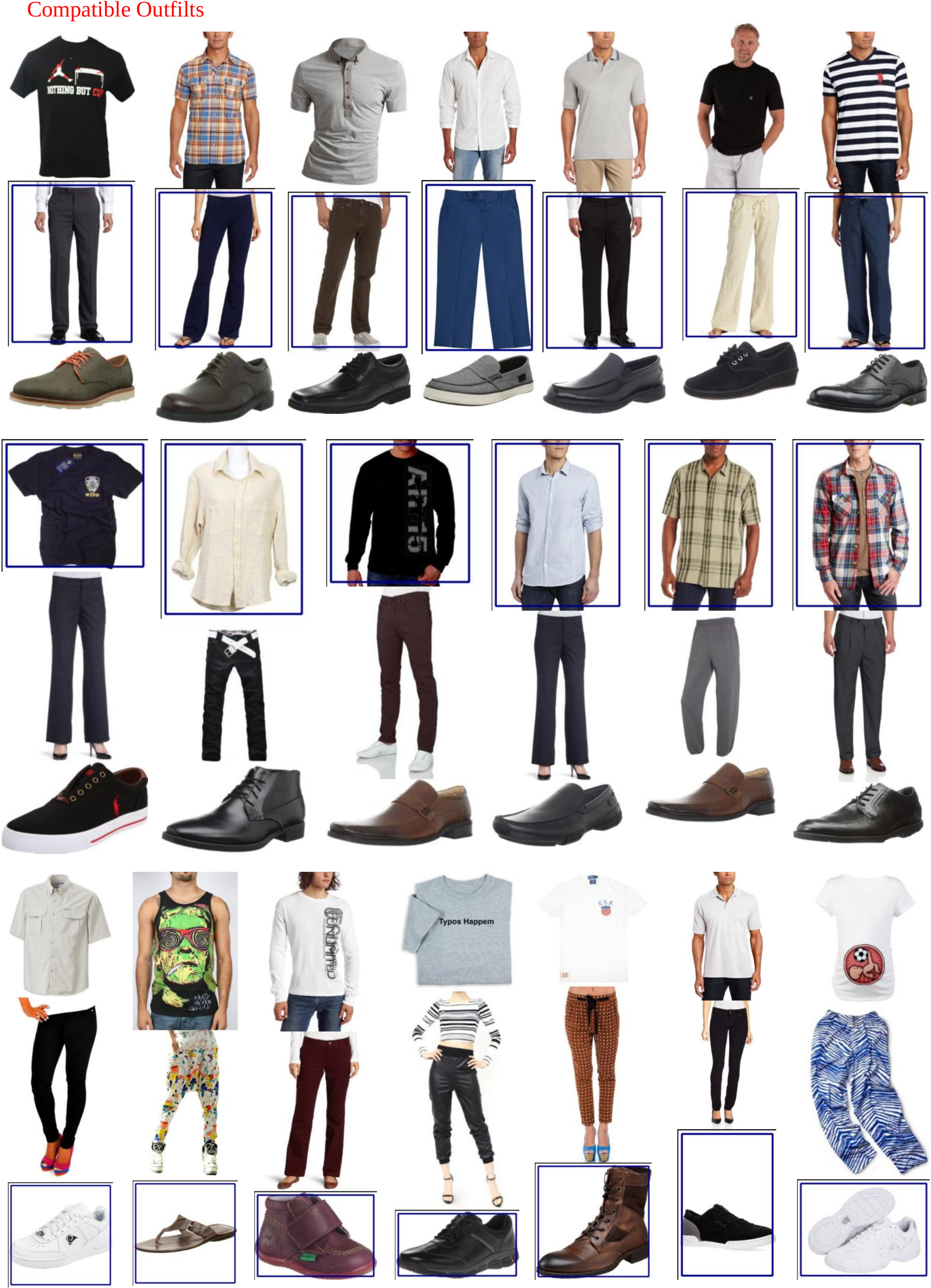}
		\caption{Outfits composed of three products from different classes predicted 
		with high compatibility based on the proposed method. Based on the query 
		product (marked in blue) compatible products of other classes are selected.}
\label{fig:generatedOutfitP3}
\end{figure}

\begin{figure}[h!]
	\centering
		\includegraphics[width=0.99\textwidth]{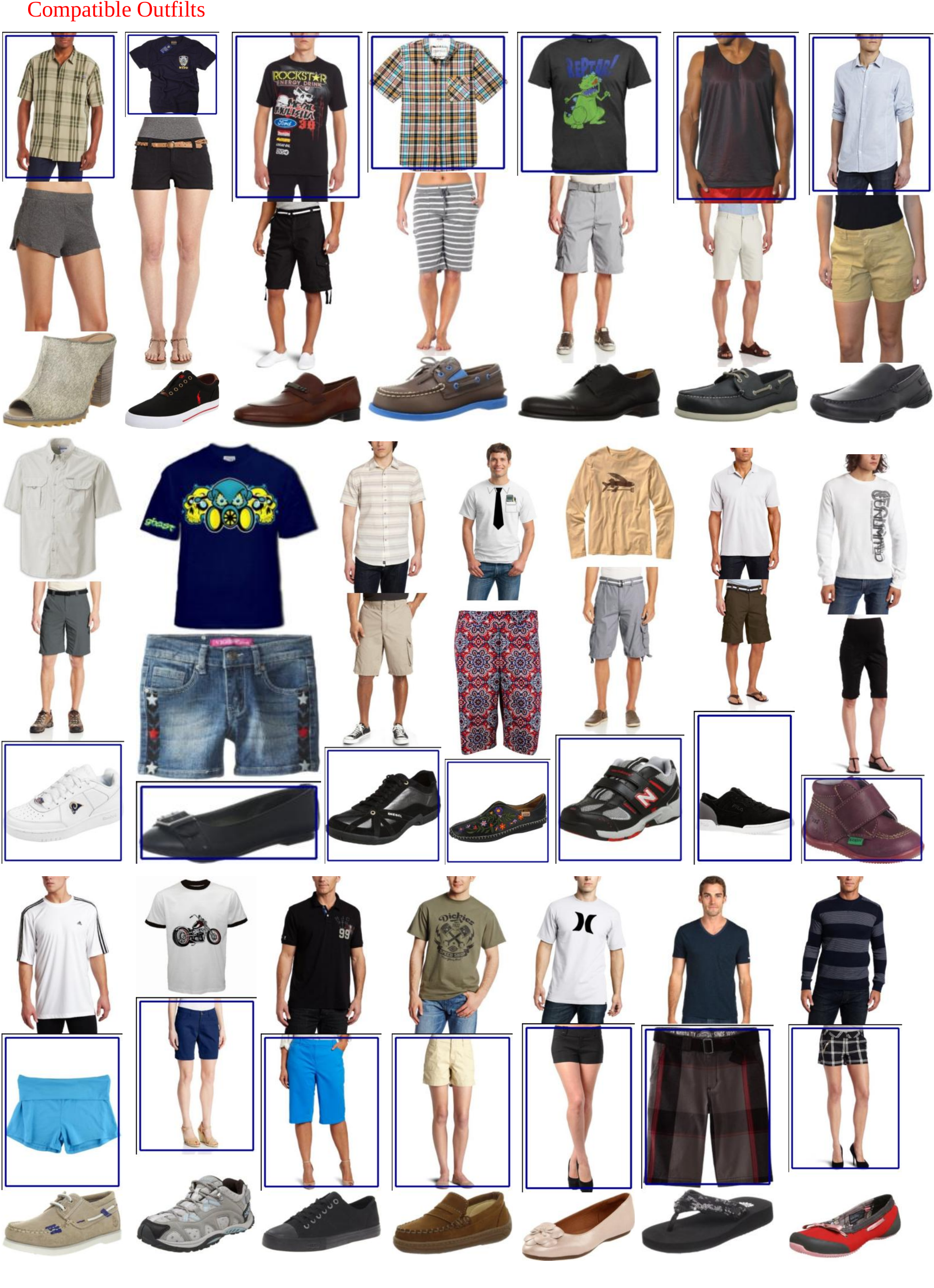}
		\caption{Outfits composed of three products from different classes predicted 
		with high compatibility based on the proposed method. Based on the query 
		product (marked in blue) compatible products of other classes are selected.}
\label{fig:generatedOutfitP4}
\end{figure}

\begin{figure}[h!]
	\centering
		\includegraphics[width=0.99\textwidth]{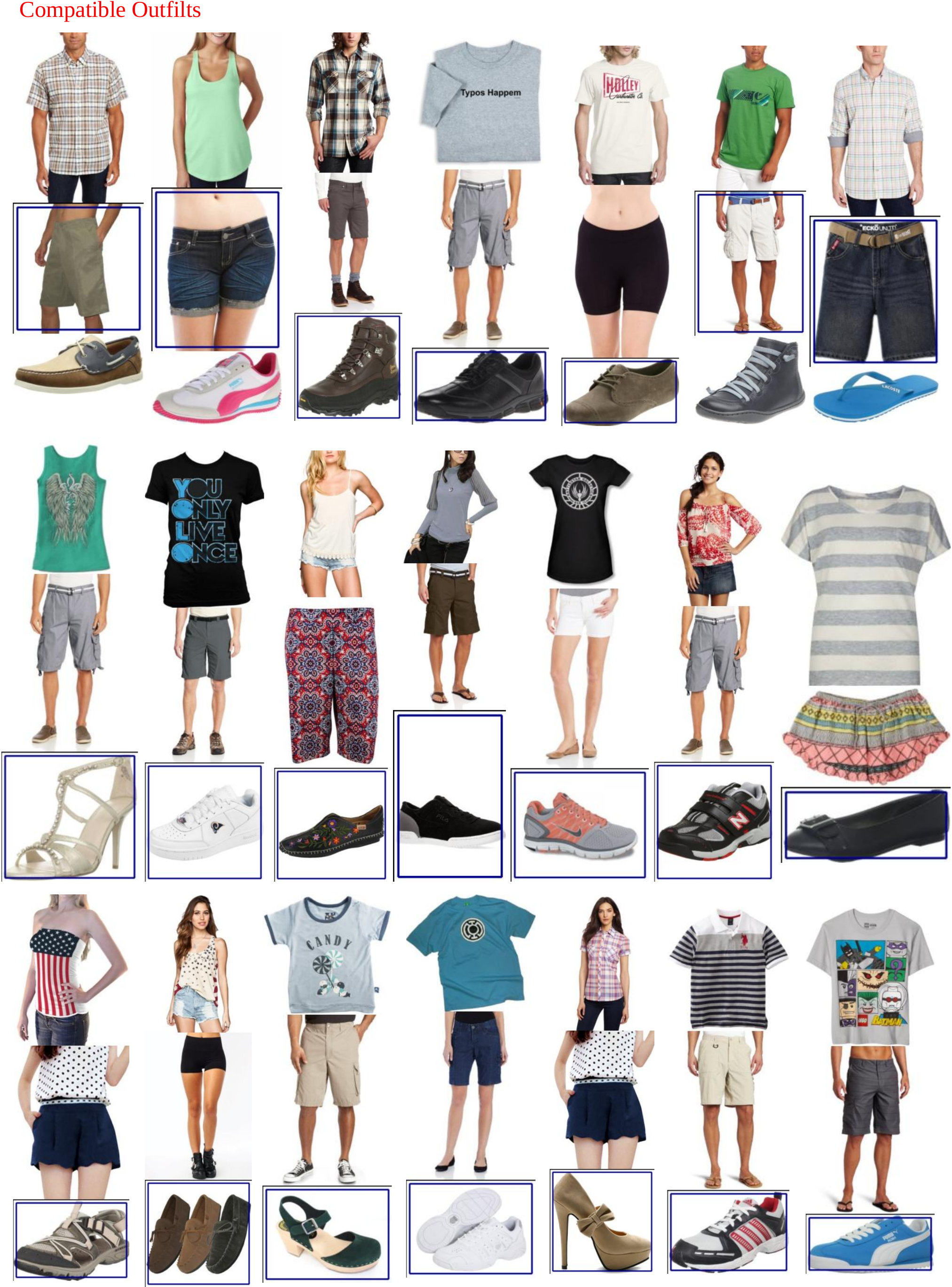}
		\caption{Outfits composed of three products from different classes predicted 
		with high compatibility based on the proposed method. Based on the query 
		product (marked in blue) compatible products of other classes are selected.}
\label{fig:generatedOutfitP5}
\end{figure}

\begin{figure}[h!]
	\centering
		\includegraphics[width=0.99\textwidth]{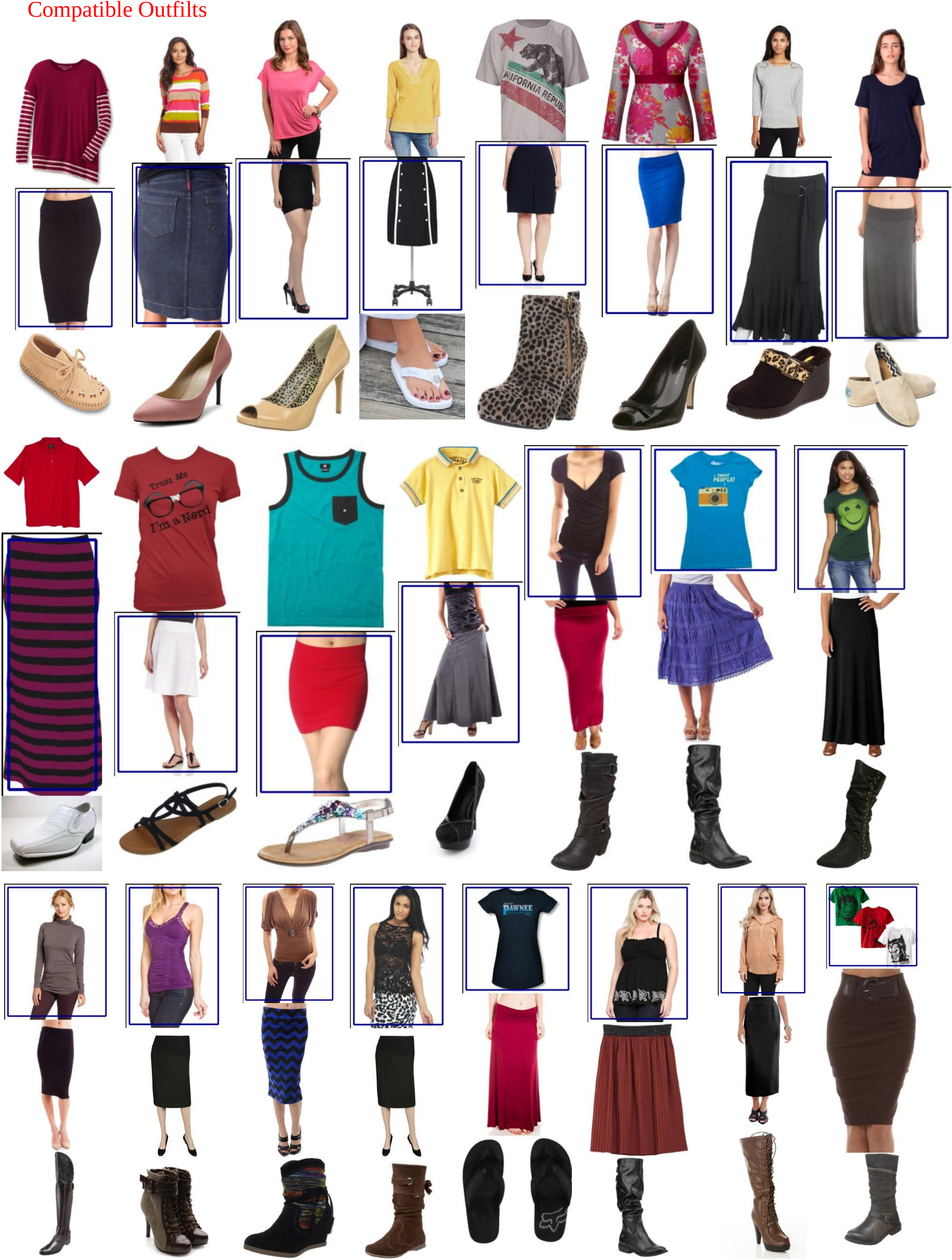}
		\caption{Outfits composed of three products from different classes predicted 
		with high compatibility based on the proposed method. Based on the query 
		product (marked in blue) compatible products of other classes are selected.}
\label{fig:generatedOutfitP6}
\end{figure}

\begin{figure}[h!]
	\centering
		\includegraphics[width=0.99\textwidth]{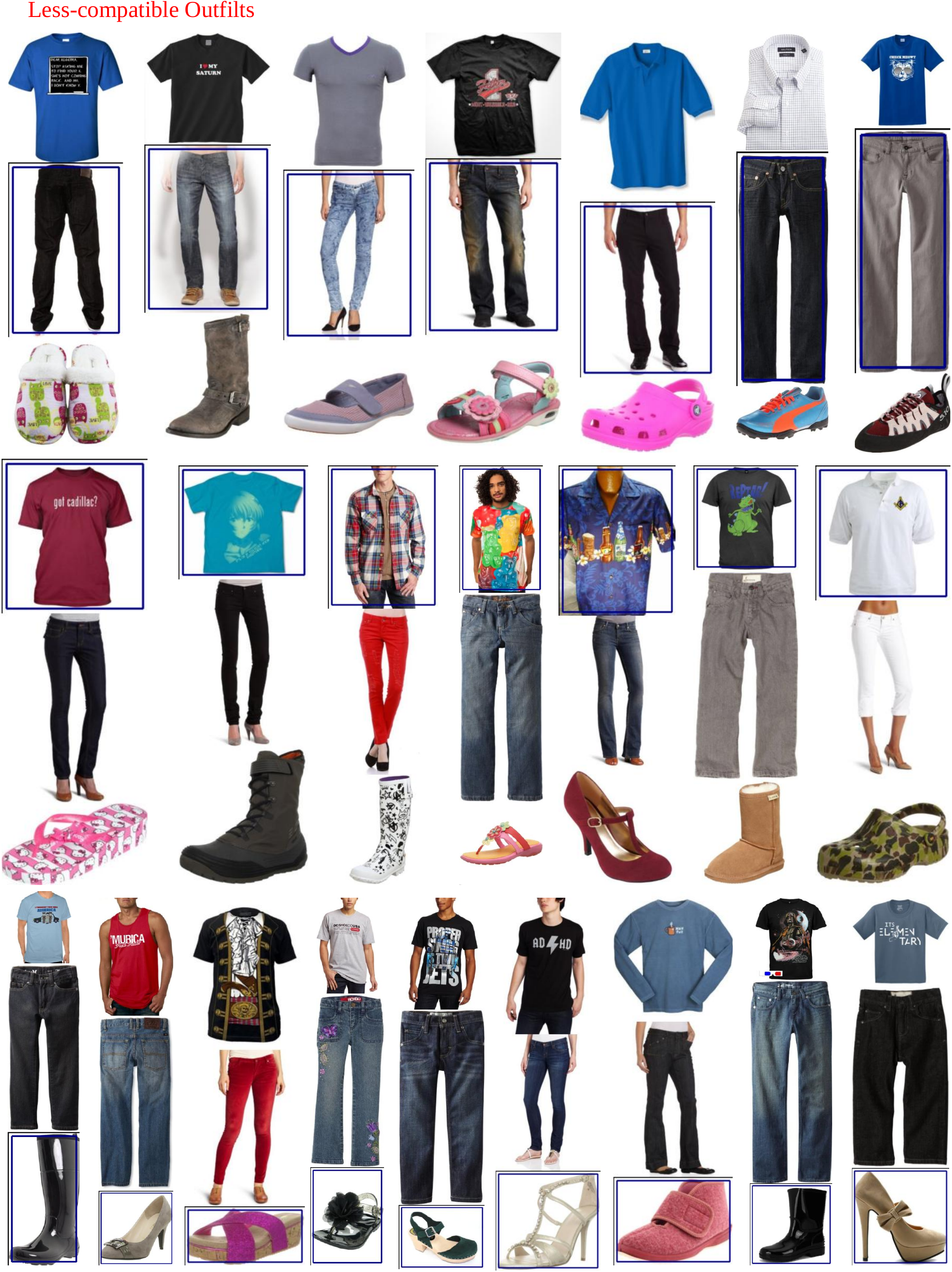}
		\caption{Outfits composed of three products from different classes predicted 
		with the \textbf{\textit lowest} compatibility score based on the proposed method. Based on the query 
		product (marked in blue) compatible products of other classes are selected.}
\label{fig:generatedOutfitP7}
\end{figure}

\begin{figure}[h!]
	\centering
		\includegraphics[width=0.99\textwidth]{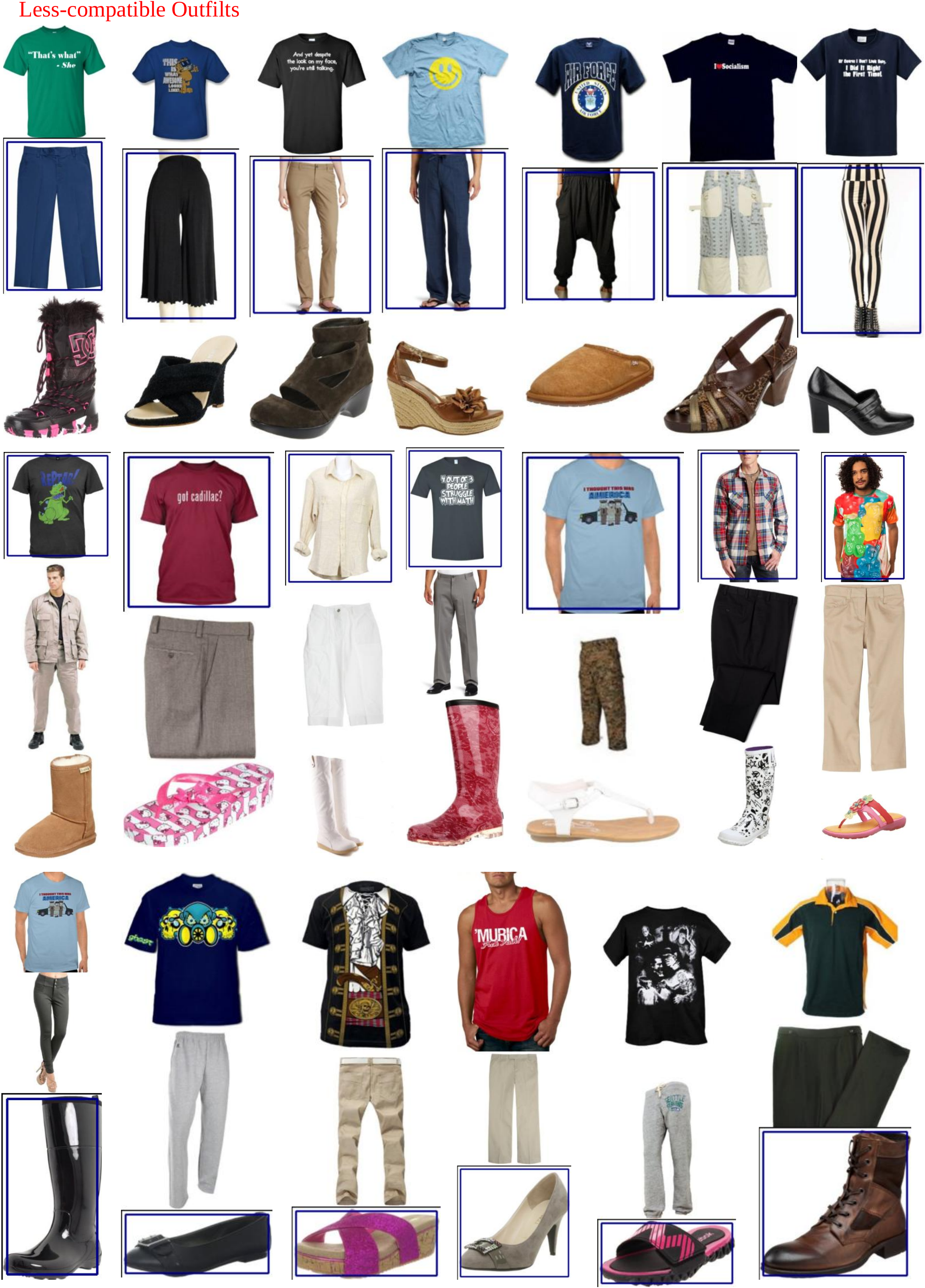}
		\caption{Outfits composed of three products from different classes predicted 
		with the \textbf{\textit lowest} compatibility score based on the proposed method. Based on the query 
		product (marked in blue) compatible products of other classes are selected.}
\label{fig:generatedOutfitP8}
\end{figure}

\begin{figure}[h!]
	\centering
		\includegraphics[width=0.99\textwidth]{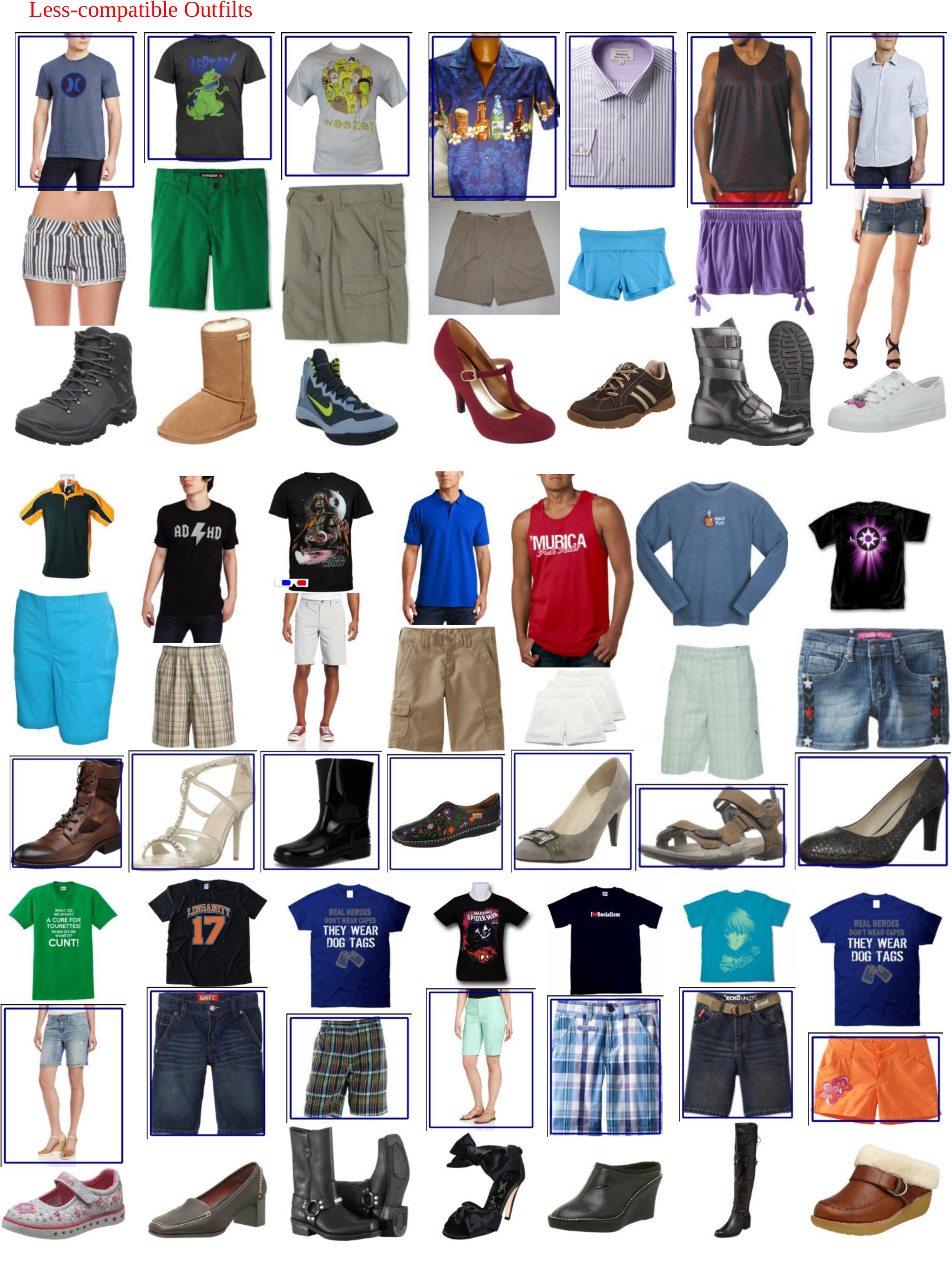}
		\caption{Outfits composed of three products from different classes predicted 
		with the \textbf{\textit lowest} compatibility score based on the proposed method. Based on the query 
		product (marked in blue) compatible products of other classes are selected.}
\label{fig:generatedOutfitP9}
\end{figure}

\begin{figure}[h!]
	\centering
		\includegraphics[width=0.99\textwidth]{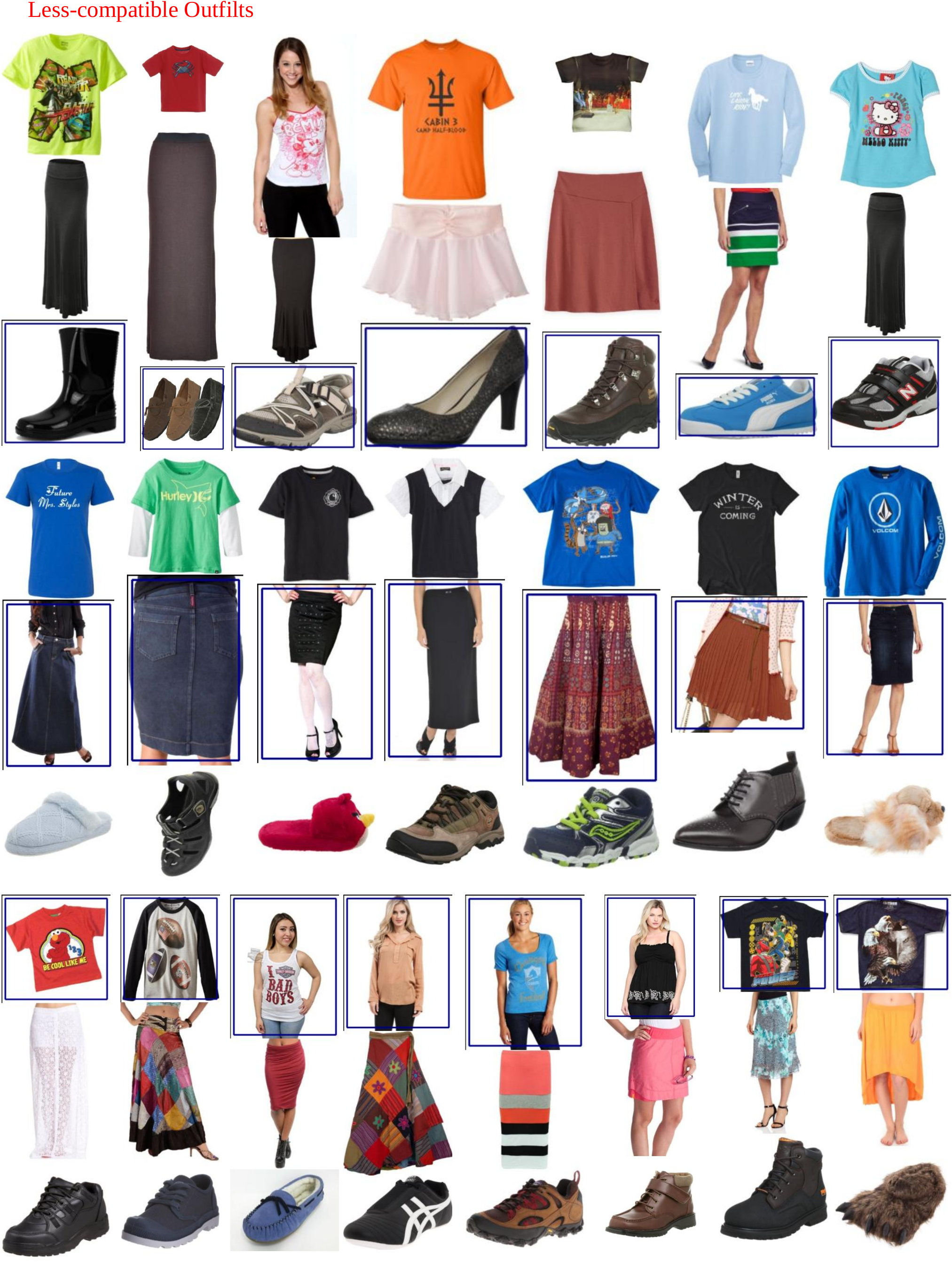}
		\caption{Outfits composed of three products from different classes predicted 
		with the \textbf{\textit lowest} compatibility score based on the proposed method. Based on the query 
		product (marked in blue) compatible products of other classes are selected.}
\label{fig:generatedOutfitP10}
\end{figure}

This Section constitutes supplementary material to the manuscript presented above.
This document extends the original manuscript in the following directions:

\begin{itemize}
 \item Figures \ref{fig:productStyleClustersP1} and \ref{fig:productStyleClustersP2}, extend Figure 3 of the previous manuscript by presenting additional examples of the extracted base-level
 visual elements (Section 4.1 of the previous manuscript). These base-level elements are extracted from the Amazon dataset \cite{McAuleyKDD15,McAuleySIGIR15} for each of the classes of interest (\textit{accessories, jeans, 
pants, shirts, shoes, shorts, skirts, and tops \& tees}). Figure~\ref{fig:generatedOutfitP1} and  \ref{fig:generatedOutfitP2} present examples of these classes clustered based on their respective visual base-level elements. Note how these base-level elements effectively describe some of the variations or 'styles' in which the objects of interest may occur.

 \vspace{0.2cm}

\item In Figures \ref{fig:generatedOutfitP1} -\ref{fig:generatedOutfitP10} we present outfits generated based on the visual compatibility predicted by the proposed method (Section 4.3 of the previous manuscript). We generate these outfits by defining a subset of three classes from which a query class is selected. Then, the visual compatibility w.r.t. to the other classes is computed for every example within each class. This procedure is performed for the examples on the image test set from the Amazon dataset \cite{McAuleyKDD15,McAuleySIGIR15} defined in \cite{VeiKovBelMcABalBel15}. In Figures \ref{fig:generatedOutfitP1}-\ref{fig:generatedOutfitP6} we show examples of generated outfits with high visual compatibility score (Section 4.3 of the previous manuscript). For reference, in Figures \ref{fig:generatedOutfitP7}-\ref{fig:generatedOutfitP10} we present \textit{less-compatible} outfits, i.e. outfits with the lowest compatibility score. Note that when generating outfits, we only measure visual compatibility 
between the query example and the examples from the other classes, one example and class at a time. At this point, no compatibility between the retrieved examples is estimated.
  
\end{itemize}


\end{document}